\documentclass[journal,10pt]{IEEEtran}
\usepackage{amsmath,amssymb}
\usepackage{algorithm}
\usepackage{algorithmic}
\usepackage{array}
\usepackage{booktabs}
\usepackage{graphicx}
\usepackage{url}
\usepackage{xcolor}
\usepackage[colorlinks=true,linkcolor=blue,urlcolor=blue,citecolor=blue,bookmarksnumbered=true]{hyperref}
\hypersetup{pdftitle={PF-Exec: Large-Scale UAV Relay Load Balancing with an Anchored Potential Direction Decider and a Per-Dimension Gain-Learning Executor},pdfauthor={Liheng Zeng, Hanzhang Jia, Yiran Li, Yi Gao, Bo Ma}}

\begin{document}

\title{PF-Exec: Large-Scale UAV Relay Load Balancing with an \\ Anchored Potential Direction Decider and a \\ Per-Dimension Gain-Learning Executor}

\author{Liheng Zeng,$^{1,\dagger}$ Hanzhang Jia,$^{1,\dagger,\ast}$ Yiran Li,$^{1}$ Yi Gao,$^{2}$ and Bo Ma$^{3,\ast}$%
\thanks{L. Zeng, H. Jia and Y. Li are with the University of Sussex.}%
\thanks{Y. Gao is with Shanghai Shuyuan Information Technology Co., Ltd.}%
\thanks{B. Ma is with Zhejiang Gongshang University.}%
\thanks{$^{\dagger}$These authors contributed equally to this work. $^{\ast}$Corresponding authors. The primary corresponding author is Hanzhang Jia (jhzaliyy@outlook.com), and the secondary corresponding author is Bo Ma (mabo@mail.zjgsu.edu.cn).}}

\markboth{IEEE Transactions on Wireless Communications, submitted for review}%
{Zeng \MakeLowercase{\textit{et al.}}: Large-Scale UAV Relay Load Balancing}

\maketitle

\begin{abstract}
Relay load balancing in large-scale UAV networks requires joint decisions on association, bandwidth, power and hovering position, while the dimension grows with the number of UAVs and each UAV sees only its local state. This paper splits the decision into two layers. An anchored potential direction decider emits a closed-form target at every instant and returns its direction only. A gain-learning executor maps a local observation into a positive gain per dimension, composes gain and target, and clips the action against the physical single-step bound; the two layers form PF-Exec. The unit gain is the default start, so the executor reduces to the pure potential policy before training. The load is first-order homogeneous in bandwidth, making the bandwidth target closed form, and the position target is the demand-weighted centroid of the served users. The system uses a two-hop access and millimetre-wave backhaul link with soft association, a 3GPP UMa-AV channel with two shadowing layers, an OU-type social-force crowd and a quadrotor physical layer. Evaluation spans all fleet sizes, two crowd scenarios, independent seeds, the published baselines in their own settings and a degraded-information line. In the highest-pressure burst setting the fixed-step comparison breaks down on part of the seeds, losing fairness and mean load at once, while PF-Exec holds fairness near one. Among the ten comparison methods only the greedy heuristic survives that setting; its load variance stays far higher. Under stale global state and position noise the fairness index stays high at both largest-scale scenarios. Flight energy is nearly identical between the learned and the fixed-step executor, and three constants are fixed once across all settings. The convergence properties of both layers, from the truncated gain iteration to the perturbed tracking bound, are proved in the companion analysis \cite{r67}.

\end{abstract}


\section{Introduction}

\subsection{Motivation}

Aerial relaying extends coverage and capacity by keeping a station on the move. A UAV hovers between the host base station and the ground users, receives over a backhaul hop and forwards over an access hop \cite{r3}, \cite{r20}. Because the position of a UAV changes the access channel, the backhaul distance and the interference it suffers at the same time, flattening the load across the fleet is a joint problem of deployment, association and resource allocation. A crowd that moves unevenly, or that gathers without notice, leaves a few UAVs in deep overload while others stay lightly loaded. A fleet total does not register that skew, and the skew has a hard consequence, since a load above five means that the UAV cannot serve its users at all. Balance is therefore an independent design objective and not a by-product of coverage maximisation.

Four requirements follow from the setting. The system must serve ground users over a moving crowd and reward balance as a first-class objective. It must scale with the number of UAVs. It must run online at second-level latency. Those requirements exclude a centralised non-convex re-solution at every step and they exclude end-to-end learning whose cost couples to the fleet size. The system of this paper meets all four. The direction is closed form and local, the execution is a small network per UAV with an observation that does not grow with the fleet, and the runtime guarantee is independent of the number of UAVs.

\subsection{Related Work}

Deployment work spans single and few UAV placement \cite{r6,r7,r8,r9,r10}, density-aware placement \cite{r8}, joint coverage \cite{r11,r12,r13,r14}, and formulations that include the backhaul \cite{r15}, \cite{r36}. Load-balancing work studies association \cite{r17,r18,r19}, load-aware aerial deployment \cite{r21,r22,r23,r24}, proportional and alpha fairness \cite{r30,r31,r32,r33}, \cite{r38,r39,r40,r41,r42}, and the delay-driven joint placement and association of \cite{r43}, which carries a complete backhaul model. Learning-based work covers deep reinforcement learning \cite{r25}, \cite{r29}, \cite{r53,r54,r55}, with a DDPG deployment agent \cite{r53}, a masked SDQN agent \cite{r54}, a mean-field method \cite{r29} and a multi-agent attention method \cite{r55}. Potential-function and centroid heuristics, virtual forces and Lloyd iterations among them, give a closed-form direction with a fixed speed \cite{r31}, \cite{r44}.

The literature splits into three families by what each one fixes and what each one leaves free, and the split explains the gap this paper addresses.

\subsubsection{Placement and Association Without a Direction}

The first family computes positions, bandwidths or assignments by solving a program over the whole fleet. The early placement work on aerial base stations \cite{r3,r20} fixes a deployment objective and search over a lattice or over a continuous box; density-aware variants \cite{r8} weight the search by the expected number of users per cell; coverage-oriented formulations \cite{r11,r12,r13,r14} add a disk model of the served area and solve a discrete covering problem; and backhaul-aware variants \cite{r15,r36} add a capacity constraint on the wireless backhaul. Load-balancing formulations \cite{r17,r18,r19,r21,r22,r23,r24} add a fairness objective on the rates and solve the resulting assignment and bandwidth program by mixed-integer programming or by a convex relaxation with rounding. The delay-driven joint placement and association of \cite{r43} is the most complete of these, since it carries a full backhaul model and solves the joint program with a successive convex approximation. The common feature is that the object computed is an optimiser for the current instance, so the whole program has to be re-solved whenever the instance moves, and the cost of the re-solution grows with the fleet.

\subsubsection{Fairness as an Indicator Rather Than a Target}

The second family studies what fairness means and how to measure it on a fleet of access points. Proportional fairness \cite{r30,r31,r32} and the alpha-fair family \cite{r33,r38,r39,r40,r41,r42} give a one-parameter family of objectives whose maximisers range from throughput maximisation to max-min fairness, and the Jain index \cite{r37} gives the scalar that is read off the resulting allocation. These papers establish the two facts this paper leans on: that the equal-load configuration is the maximiser of the concave members of the family, and that the Jain index is a monotone transform of those objectives in the neighbourhood of the equal-load point. What they do not supply is a closed-form map from a measured state to the allocation that reaches that point, which is what a fleet needs when the state moves on a control period.

\subsubsection{Learning the Allocation End to End}

The third family replaces the program by a parametric policy. Deep reinforcement learning for deployment \cite{r25,r29,r53,r54,r55} trains an agent whose input is the state and whose output is a position, a power or an association, with a DDPG agent \cite{r53}, a masked structured deep Q-network \cite{r54}, a mean-field relaxation of the multi-agent interaction \cite{r29} and an attention-based multi-agent critic \cite{r55}. The strength of the approach is that it handles objectives that are hard to write as a program. The cost is that the output of the agent is a number whose relation to the underlying allocation is not available, so the behaviour of the fleet under an unseen pressure cannot be predicted from the policy, and the amount of state that has to be evaluated centrally grows with the number of agents.

\subsubsection{Closed-Form Directions With a Fixed Speed}

A fourth line works from a potential: it solves the decoupled subproblems in closed form and walks the resulting direction at a fixed speed. Virtual-force methods and Lloyd-type iterations belong here, and so do the potential-based methods of \cite{r31} and \cite{r44}. The direction is available in closed form and the fleet needs no central solver, which is why these methods scale. What these methods lack is speed, not direction: the same distance is covered per control period in every regime, so the setting that is stable when the fleet is lightly loaded is either too slow when the load concentrates or unstable when the local curvature is high.

The difference of this paper lies in the decomposition. Conventional potential-based approaches provide a closed-form direction together with a fixed speed, end-to-end learning gives no direction at all, and this paper takes the direction in closed form and learns only the speed. Two facts supply the direction. In the access-limited region the load is first-order homogeneous in bandwidth, which yields the bandwidth target in closed form together with its feasibility bound. In position the objective is a weighted centroid, which yields the position target in closed form. One fact supplies the need for learning. The ill-conditioning gain grows as the spectral sum falls or the carried demand leaves the mid range, so a single fixed step cannot keep one contraction condition across the pressure range, as Proposition~7 of the companion paper \cite{r67} shows.

\subsection{Contributions}

\begin{itemize}
\item A complete system for relay load balancing. The end-to-end available rate is the smaller of the access rate and the backhaul capacity, association is soft through a damped softmax, the channel follows the 3GPP UMa-AV model with a spatially correlated shadowing layer and a runtime shadow, the crowd moves under an OU-type social force, and the airframe is simulated on a quadrotor physics layer. The load is demand over end-to-end available rate, and the overload and dead thresholds are explicit.
\item An anchored potential direction decider. The direction ranks load under the equal-weight complete-graph potential of a logarithmic knee view, which preserves order over the whole load domain and keeps a strictly positive gradient in deep overload. The bandwidth target is the demand scale over the anchor, where the demand scale $c_i=u_ib_i$ is measured to be independent of bandwidth with a drift of $10^{-16}$, and the target degenerates into a proportional split when the budget binds.
\item A per-dimension gain-learning executor. A 22-dimensional local observation feeds a projected residual block that emits five positive gains under a hard bound of $[0.50,2.01]$, trained by TD3. The unit gain at initialisation is exactly the pure potential policy, so the executor improves through learning and never passes through an unsafe intermediate. Two further executors with the same direction, one at a constant maximal speed and one at a fixed step, separate the contribution of the direction from the contribution of the speed.
\item A full-scale evaluation of 602 runs. The runs cover three, twelve, thirty, sixty and one hundred UAVs, the flow and burst scenarios, three seeds, ten comparison methods, four native settings of published baselines and a real information-degradation line. Every grid of every seed is reported for the blocks that carry three seeds, with the minimum and the range made explicit.
\end{itemize}

\noindent\textbf{Companion analysis.} The convergence properties are proved in \cite{r67}, which establishes the homogeneous decomposition and the anchored closed form, the complete dynamics of the one-dimensional truncated gain, and the step-size ceiling. That paper also reports that the spectral sufficient condition fails on sixty of sixty measured states, so convergence here rests on the gain bound together with truncation, and no spectral certificate is required. This paper measures behaviour, and the companion paper explains it.

\section{System Model}

\subsection{Network and Two Hops}

\begin{figure}[!t]
\centering
\includegraphics[width=\columnwidth]{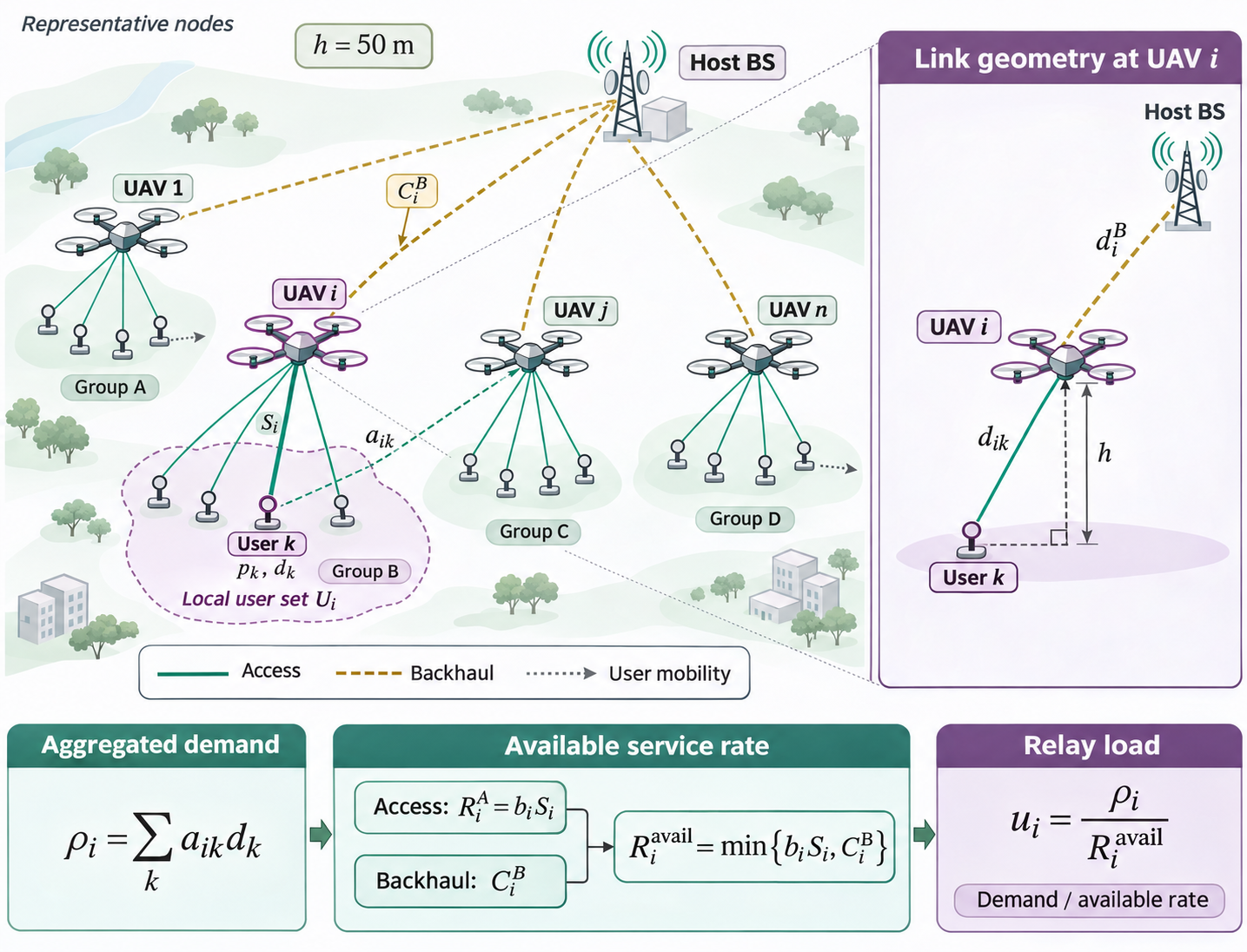}
\caption{System model. A host base station sits at the origin and hovers serve ground users over an access hop, drawn solid, while the backhaul hop is drawn dashed. Every UAV keeps a fixed hovering height above a horizontal position inside the field.}
\label{fig:system}
\end{figure}

Consider $n$ UAVs, $K$ users and one host base station at the origin, as Fig.~\ref{fig:system} shows. Every UAV hovers at a fixed height $h=50$ m over a horizontal position inside the field, receives over a backhaul hop at 28 GHz and forwards over an access hop at 2 GHz. The end-to-end capacity of a UAV is the smaller of its total access rate and its backhaul capacity. When the backhaul binds, the rates of its associated users are scaled down by a common factor, which keeps the association structure of \cite{r43}. The environment step is 0.1 s, an episode is 2000 steps, that is 200 s, and the evaluation extends every trained policy to 4000 steps without resetting.

\subsection{Problem Statement}

Let $b_i$ be the bandwidth of UAV $i$, $\pi_i$ its price signal, $p_i$ its power multiplier and $\xi_i=(\xi_{i1},\xi_{i2})$ its horizontal position, and let $X=[x_{ik}]$ be the association matrix that the damped softmax produces from the channel and from $(b,\pi)$. The per-user shares follow from the demand proportions and the load $u_i=\rho_i/R_i^{\mathrm{avail}}$ follows from the access rate and the backhaul cap. The decision problem asks for a flat load profile, and it is written
\begin{equation}
\begin{aligned}
\min_{b,\pi,p,\xi}\ \ & \frac{1}{n}\sum_{i=1}^{n}\bigl(\hat u_i-\bar{\hat u}\bigr)^2\\
\text{s.t.}\ \ & \textstyle\sum_i b_i\le B,\quad b_i\in[10^{-3},B],\quad \pi_i\in[-3,3],\\
& p_i\in[0.5,2],\quad \xi_i\in\mathcal F,\quad X=\mathrm{Assoc}(b,\pi;G),
\end{aligned}
\label{eq:problem}
\end{equation}
with $\mathcal F$ the field rectangle, $G$ the channel table and $\bar{\hat u}$ the fleet mean of the view. The objective is the variance of the viewed load, which is the potential of Section~III, and the association map enters as an equality constraint that is set by the same decision variables.

The program of (\ref{eq:problem}) couples every UAV to every other through the association and through the bandwidth budget, and the coupling does not shrink with $n$. Three structural facts remove the coupling in closed form. The load is first-order homogeneous in bandwidth in the access-limited region, so the equal-load bandwidth target is a ratio of the demand scale to the load anchor, and the budget enters only as a common level. The position target is separable per coordinate, so the constrained minimiser of the position part is the demand-weighted centroid of the served users. The price signal enters through a subgradient that is a difference of viewed loads, so the price part needs one step against the local load only. The decider therefore never solves (\ref{eq:problem}); it emits the direction that (\ref{eq:problem}) points along at the current state, and the executor decides how far to move along that direction. The two bounds that carry over from (\ref{eq:problem}) into the implementation are the bandwidth budget and the physical single-step displacement.

\subsection{Backhaul Hop}

The backhaul link follows
\begin{equation}
L_i^{\mathrm B}=61.4+20\log_{10}d_i^{\mathrm B},\qquad
d_i^{\mathrm B}=\bigl(x_i^2+y_i^2+h^2\bigr)^{1/2},
\label{eq:backhaul}
\end{equation}
and the backhaul capacity is
\begin{equation}
C_i^{\mathrm B}=B^{\mathrm B}\log_2\Bigl(1+\frac{p^{\mathrm B}g_i^{\mathrm B}}{I^{\mathrm B}}\Bigr),\quad
g=10^{-L/10},\quad I^{\mathrm B}=N_0^{\mathrm B}B^{\mathrm B}.
\end{equation}
The backhaul bandwidth $B^{\mathrm B}=100$ MHz is orthogonal across UAVs, the backhaul power is $p^{\mathrm B}=43$ dBm and the noise density is $N_0^{\mathrm B}=-174$ dBm/Hz. Small-scale fading is taken at its statistical mean of one, so the backhaul capacity depends on distance alone. Over the field of the base setting the capacity runs from 522 Mbps at a thousand metres to 913 Mbps at two hundred and fifty metres, and stays fifteen to two hundred and twenty six times the access rate in the normal regime.

\subsection{Access Hop}

The access channel follows TR~36.777 for aerial users \cite{r58} and TR~38.901 for shadowing parameters \cite{r59}. The path loss is baked offline on a ten-metre grid,
\begin{equation}
\begin{split}
\mathrm{PL}_{\mathrm{LoS}}=\max\Bigl\{\mathrm{FSPL},\,
&30.9+(22.25-0.5\log_{10}z)\log_{10}d_{3D}\\
&+20\log_{10}f\Bigr\},
\end{split}
\end{equation}
\begin{equation}
\begin{split}
\mathrm{PL}_{\mathrm{NLoS}}=\max\bigl\{\mathrm{PL}_{\mathrm{LoS}},
&32.4+(43.2-7.6\log_{10}z)\log_{10}d_{3D}\\
&+20\log_{10}f\bigr\},
\end{split}
\end{equation}
with $z=50$ m and $f=2$ GHz. The line-of-sight state is drawn from the elevation-dependent probability with a fixed seed. The spatially correlated shadowing is generated by the QuaDRiGa sum-of-sinusoids engine \cite{r66} at $\sigma=4$ dB over 37 m for the line-of-sight case and $\sigma=6$ dB over 50 m otherwise, both taken from the standard configuration. Online the propagation term is fitted analytically as
\begin{equation}
G_{ik}=C_0[k]-10\gamma\log_{10}\bigl(d_{ik}+50\bigr)+s_{ik},
\label{eq:fit}
\end{equation}
where $C_0[k]$ is a per-user intercept, $\gamma$ is the fleet median and $s_{ik}$ is a zero-mean shadow whose standard deviation is drawn uniformly from $[0,13.2]$ dB at the start of every episode and refreshed every hundred steps, which matches a coherence distance of fifty metres. The runtime shadow and the baked shadow are independent layers.

\subsection{Association, Rate and Load}

Association is soft, with $x_{ik}\ge0$ and $\sum_ix_{ik}=1$. The initial weights are proportional to $\exp(G_{\cdot k}/10)$. The weights follow a damped softmax
\begin{equation}
\begin{split}
x^+&=\Bigl(1-\frac1{\tau_u}\Bigr)x+\frac1{\tau_u}\operatorname{softmax}_i\bigl(z_{ik}\bigr),\\
z_{ik}&=\alpha r_{ik}+\kappa\frac{R_i^{\mathrm{avail}}}{\rho_i+\varepsilon}+\pi_i,
\end{split}
\label{eq:assoc}
\end{equation}
with $\tau_u=5$, $\alpha=1.0$, $\kappa=0.5$ and $\varepsilon=0.1$, where $\pi_i\in[-3,3]$ is a dual price per UAV that balances load across the fleet. The signal-to-interference-plus-noise ratio is
\begin{equation}
\mathrm{SINR}_{ik}=\frac{p_ig_{ik}}{N_0+\sum_{j\ne i}x_{jk}p_jg_{jk}},\qquad p_i=P\tilde p_i,
\end{equation}
with $P=24$ dBm, a power multiplier $\tilde p_i\in[0.5,2]$ and noise $N_0=-104$ dBm, and the weighted sum in the denominator is the soft-association interference. Bandwidth is split in two stages. The fleet total obeys $\sum_ib_i\le B$ and is scaled back proportionally when the sum exceeds it, and inside a UAV the share of a user is $b_{ik}=x_{ik}d_kb_i/\sum_jx_{jk}d_j$. The rates and the load are
\begin{equation}
\begin{split}
R_{ik}&=b_{ik}\log_2(1+\mathrm{SINR}_{ik}),\quad R_i^{\mathrm A}=\sum_kx_{ik}R_{ik},\\
\rho_i&=\sum_kx_{ik}d_k,
\end{split}
\end{equation}
\begin{equation}
R_i^{\mathrm{avail}}=\min\{R_i^{\mathrm A},\,C_i^{\mathrm B}\},\qquad u_i=\frac{\rho_i}{R_i^{\mathrm{avail}}}.
\label{eq:load}
\end{equation}
A load above one means that the demand of a UAV cannot be served, and a load above five makes the UAV dead, since no user obtains a usable rate. Balance is reported by the Jain fairness index \cite{r37}, $J=\bar u^2/(\bar u^2+\operatorname{var}_u)$. The denominator of the load is the quantity that makes the two hops interact: $R_i^{\mathrm{avail}}$ is the smaller of the access rate and the backhaul rate, so a UAV carries the demand of its users at whichever hop is tighter and the load responds to a change in either. Two consequences matter for the direction. A UAV whose backhaul binds gains nothing from extra bandwidth, so bandwidth sent to it is wasted when the fleet is short of backhaul, and a UAV whose access hop binds gains rate in proportion to bandwidth, which is what makes the load first-order homogeneous there and gives the closed-form target. The frontier between the two hops moves as the fleet spreads out, since the backhaul distance of a UAV grows with its displacement while the access distance to its users falls. The direction follows that frontier without being told which hop is active, because the load ratio already contains the minimum.

The serving hop also explains the shape of the load under a gathering. When the crowd concentrates into two exits, the access distance of a UAV falls for the users near an exit and rises for the rest, so the per-user spectral efficiency spreads. A UAV that keeps its assignment spreads its rate over a wider range, its available rate falls at the access hop, and its load rises. The bandwidth target responds by sending it a share proportional to its demand scale, which is the mechanism that holds the fleet on the anchor for as long as the budget allows. A guard keeps the denominator at $10^{-6}$ Mbps, below which the load is read as zero.

\subsection{Crowd and Demand}

The crowd is a dynamic system, driven by an OU-type social force integrated at every environment step,
\begin{equation}
\dot v_i=\frac{v_i^{\mathrm{des}}e_i-v_i}{\tau_c}+\sum_jf_{ij}+\sum_Bf_{iB}+f_i^{\mathrm{grp}}.
\label{eq:crowd}
\end{equation}
The term $\tau_c=0.5$ s is the velocity relaxation time, and the desired term is exactly the drift of an Ornstein-Uhlenbeck process \cite{r62}, which matches the measured power spectral density of pedestrian velocity, proportional to $1/f^2$ with exponentially decaying autocorrelation \cite{r63}. The social and obstacle forces take the original form of Helbing and Moln\'ar \cite{r61}, and the group term takes the extension of Moussa\"id et al.\ \cite{r64}. The implementation calls a BSD-3 licensed social-force engine at formula level with the constants of the sources, namely social force factor 2.1, obstacle force factor 10 with $\sigma=0.2$ m, $\lambda=2.0$, $\gamma=0.35$, $n=2$, $\varepsilon=0.005$, $n'=3$ and an agent radius of 0.35 m, and its step equals the environment step of 0.1 s. A slow group and a fast group mix deterministically by index, at desired speeds 1.39 and 0.42 m/s, and roam around their initial hot spots. The flow scenario keeps the roam throughout. The burst scenario follows the flow for the first 600 steps and then gathers the whole crowd, which moves at 2.78 m/s towards two exogenous exits placed at the two ends of the longer side, inset by ten per cent, and dwells within half of the hot-spot radius.

Demand follows $d_k=\lvert\mathcal N(0.5,0.15)\rvert$ truncated to $[0.1,1]$ and multiplied by a demand scale. Demand follows the person, so the requirement of a user does not change in time. The scale is set once by anchoring a uniform crowd of $n=12$ at the load anchor under zero action, and the same setting is used at every scale under the density-invariant convention, so no scenario carries its own scale.

\subsection{Physical Layer}

The airframe reproduces the dynamics of a small quadrotor in the CF2X class of the simulator. Its mass is 27 g, and physics runs at 120 Hz with a PID inner loop at 40 Hz, so that one environment step of 0.1 s contains twelve physics ticks and four control updates. Height is fixed at 50 m and the single-step displacement is capped at 0.5 m against a nominal 5 m/s, which measures 3.6 to 4.4 m/s in practice. The action is $\Delta_i=[\Delta b_i,\Delta\pi_i,\Delta\tilde p_i,\Delta x_i,\Delta y_i]$ with single-step bounds
\begin{equation}
U=[5\ \mathrm{MHz},\ 0.5,\ 0.1,\ 0.5\ \mathrm{m},\ 0.5\ \mathrm{m}],
\label{eq:actionlim}
\end{equation}
and the state domain is $b_i\in[10^{-3},B]$, $\pi_i\in[-3,3]$ and $\tilde p_i\in[0.5,2]$. Flight energy is accumulated per UAV as the cubic propeller power, $\sum k_f\cdot\mathrm{rpm}^3\cdot dt$, and is reported as an indicator only. The neighbourhood radius is 250 m and the neighbourhood graph is rebuilt every step from the true distances. Two slow fleet-level scalars are broadcast every 0.5 s, namely the sum of demand scales and the mean price, both of $O(1)$ magnitude; the total budget is a fixed public parameter and is not broadcast. The parameter setting is collected in Table~\ref{tab:params}.

\begin{table}[!t]
\centering
\caption{Simulation parameters, base setting at $n=12$.}
\label{tab:params}
\footnotesize
\setlength{\tabcolsep}{4pt}
\begin{tabular}{@{}>{\raggedright\arraybackslash}p{0.30\columnwidth}>{\raggedright\arraybackslash}p{0.66\columnwidth}@{}}
\toprule
Category & Value \\
\midrule
field & $600\times450$ m, height 50 m \\
scale & 3, 12, 30, 60, 100, with side $\propto\sqrt{n/12}$ and users, bandwidth $\propto n/12$ \\
crowd & slow 0.42 m/s, fast 1.39 m/s, burst 2.78 m/s, step 0.1 s \\
demand & $\lvert\mathcal N(0.5,0.15)\rvert$ truncated to $[0.1,1]$, anchor 0.8 \\
access & 2 GHz UMa-AV, 10 m grid, shadow 4 dB over 37 m and 6 dB over 50 m, runtime $U[0,13.2]$ dB every hundred steps \\
backhaul & 28 GHz, 100 MHz, 43 dBm, $-174$ dBm/Hz \\
power and noise & 24 dBm, $\tilde p\in[0.5,2]$, $-104$ dBm \\
bandwidth & 100 MHz in total, 8.33 MHz per UAV \\
association & $\tau_u=5$, $\alpha=1$, $\kappa=0.5$, $\varepsilon=0.1$, $\pi\in[-3,3]$ \\
action & five dimensions, bounds of (\ref{eq:actionlim}) \\
physical layer & CF2X 27 g, physics 120 Hz, control 40 Hz \\
episode & 2000 steps, evaluation at 2000, 3000 and 4000 \\
\bottomrule
\end{tabular}
\end{table}

\section{Direction Decider and Gain Executor}

\subsection{Load View and Potential}

The fleet potential is the equal-weight complete-graph form
\begin{equation}
\Phi(\hat u)=\frac n2\sum_i\bigl(\hat u_i-\bar{\hat u}\bigr)^2=\frac12\sum_{i<j}\bigl(\hat u_i-\hat u_j\bigr)^2,
\label{eq:potential}
\end{equation}
where the view is the logarithmic knee
\begin{equation}
\begin{split}
m(u)&=\begin{cases}u,&u\le3,\\[2pt] 3+3\ln\bigl(1+\tfrac{u-3}{3}\bigr),&u>3,\end{cases}\\
m'(u)&=\frac{1}{1+\max\{u-3,0\}/3}\in(0,1].
\end{split}
\label{eq:view}
\end{equation}
The view preserves order strictly over the whole load domain, keeps a strictly positive gradient everywhere, and preserves the ordering of deep overload, which is the property that a general truncation such as $\mathrm{clip}(u,0,3)$ or $\mathrm{clip}(u,0,30)$ only holds while the operating point stays outside the clipped region. In this system the overload points sit deep inside that region, and a truncation there annihilates the gradient, as all seven states of an earlier measurement showed. The view is shared by the first observation coordinate, the neighbourhood statistic, the load term of the reward and the potential, while every statistic reported to the outside world keeps the raw load.

\subsection{Closed-Form Targets}

In the access-limited region the access rate is first-order homogeneous in bandwidth,
\begin{equation}
R_i^{\mathrm A}=b_iS_i,\qquad S_i=\sum_k\frac{x_{ik}^2d_k}{\rho_i}r_{ik},
\label{eq:si}
\end{equation}
so the demand scale $c_i=u_ib_i=\rho_i/S_i$ does not depend on $b_i$, with a measured drift of $2.7\times10^{-16}$ against Lemma~1 of \cite{r67}. The bandwidth target is
\begin{equation}
b_i^\ast=\frac{c_i}{u^\ast},\qquad u^\ast=0.8,
\label{eq:btarget}
\end{equation}
and when $\sum_jb_j^\ast>B$ the target is scaled down uniformly until the budget is met, which leaves the direction proportional to $c$. The equal-load point is feasible if and only if $\mu\ge\mu_{\mathrm{low}}=\max\{\sum_ic_i/B,\ \max_ic_i^{\mathrm B}\}$, by Proposition~4 of \cite{r67}. The price update is
\begin{equation}
\pi_i^+=\pi_i+\bigl(m(u^\ast)-\hat u_i\bigr),
\label{eq:price}
\end{equation}
taken after removing the fleet mean and truncated to $[-3,3]$, which turns the price into the dual variable of the balance constraint. Power takes the target $\tilde p^\ast=1$ under the training reward, whose efficiency term saturates above the normal band. The position target is the demand-weighted centroid of the users a UAV serves, where a user belongs to the UAV with the largest association weight, in the manner of a Lloyd iteration.

The centroid is computed from information-limited positions, sampled every 0.5 s with ten metres of Gaussian noise and held between samples. The whole direction therefore requires no high-frequency global information.

\subsection{Direction Decider}

Algorithm~\ref{alg:dir} states the decider, run per UAV at every environment step. It reads the local state, the position and the associated users, and it emits a direction package. Every target comes from the closed forms of the preceding subsection, and the package carries the sign of each scalar dimension, the unit vector of the position dimension and the residual distance normalised by the single-step bound. The field carries no parameters. Fleet communication is limited to two slow scalars.

\begin{algorithm}[!t]
\caption{Direction decider, per UAV, per environment step}
\label{alg:dir}
\footnotesize
\begin{algorithmic}[1]
\STATE read $u_i$, $b_i$, $\pi_i$, $\tilde p_i$, position, served set and demand
\STATE $c_i\leftarrow u_ib_i$ and $b_i^\ast\leftarrow c_i/u^\ast$
\IF{$\sum_jb_j^\ast>B$}
\STATE $b_i^\ast\leftarrow b_i^\ast\cdot B/\sum_jb_j^\ast$ using the broadcast $\sum_jc_j$
\ENDIF
\STATE $\pi_i^+\leftarrow\operatorname{clip}\bigl(\pi_i+\bigl(m(u^\ast)-m(u_i)\bigr)-\bar\pi,\ \pm3\bigr)$
\STATE $\tilde p_i^\ast\leftarrow1$
\STATE $x_i^\ast\leftarrow$ demand-weighted centroid of the users whose largest weight is at $i$, from the 0.5 s held sample
\STATE emit $\operatorname{sign}(b^\ast-b)$, $\operatorname{sign}(\pi^+-\pi)$, $\operatorname{sign}(\tilde p^\ast-\tilde p)$, the unit vector towards $x_i^\ast$, and $|x^\ast-x|/U$
\end{algorithmic}
\end{algorithm}

\subsection{Three Executors}

The three executors share one direction and differ only in speed.

The constant-speed executor sets the action to the direction times the single-step bound, so its speed is always maximal. It carries no training, and it measures the value of the direction alone.

The fixed-step executor multiplies the bandwidth dimension by a relaxation of 0.2, steps the price at a constant 0.05, and drives power and position to their bounds. It carries no training either, and it is a strong fixed-gain counterpart of the proposed executor, since its four constants were tuned grid by grid.

The three families span the range of admissible speed laws, and each one answers a different question. The constant-speed family asks whether the direction alone is enough, and it settles the floor of what the decomposition contributes without any adaptation. The fixed-step family asks whether a single gain, chosen with knowledge of the grids, can hold the anchor everywhere, and it isolates the value of adaptation from the value of the decomposition. The learned family asks what a gain that depends on the local state can do where a constant cannot, and it carries the only trained parameters of the proposed method. Reading the three together separates the contribution of the closed-form direction from the contribution of the learned speed, which a comparison against external baselines alone cannot do.

The learned executor observes a 22-dimensional vector, made of twelve local environment quantities, five direction components and five normalised residual distances, and the dimension does not grow with the fleet. The network is a projected residual block, $22$--$16$--$32$--$64$--$32$--$16$--$5$, with layer normalisation and GELU activations. The head emits a positive gain per dimension through
\begin{equation}
\eta_i=\exp\bigl(0.7\tanh(\cdot)\bigr)\in[0.50,\,2.01],
\label{eq:gain}
\end{equation}
and the action composes the gain with the target,
\begin{equation}
\Delta_i=\mathrm{clip}\bigl(\eta_i\odot(X_i^\ast-X_i),\ \pm U\bigr),
\label{eq:compose}
\end{equation}
with the bandwidth and price dimensions multiplied by the relaxation of 0.2. The upper end is set above the critical gain of two on purpose, because a physical actuator carries gain error and overshoot rather than a nominal gain that stays below the threshold; the setting reproduces that non-ideal actuation and measures what the algorithm does when the critical state is actually present. The output layer starts at zero, so the gain starts at one and the executor starts as the pure potential policy.

Safety is structural and built into the action space. The hard bound of (\ref{eq:gain}) keeps every gain inside the range of Theorem~1 of \cite{r67}, which gives geometric convergence below the critical gain of two, bounded two-state oscillation at that critical gain, a family of two-state cycles of amplitude at most $(1-1/\gamma)h<h$ above it, and boundedness for every positive gain. The action clip of (\ref{eq:compose}) then bounds the displacement by the physical limit at every step.

\begin{figure}[!t]
\centering
\includegraphics[width=\columnwidth]{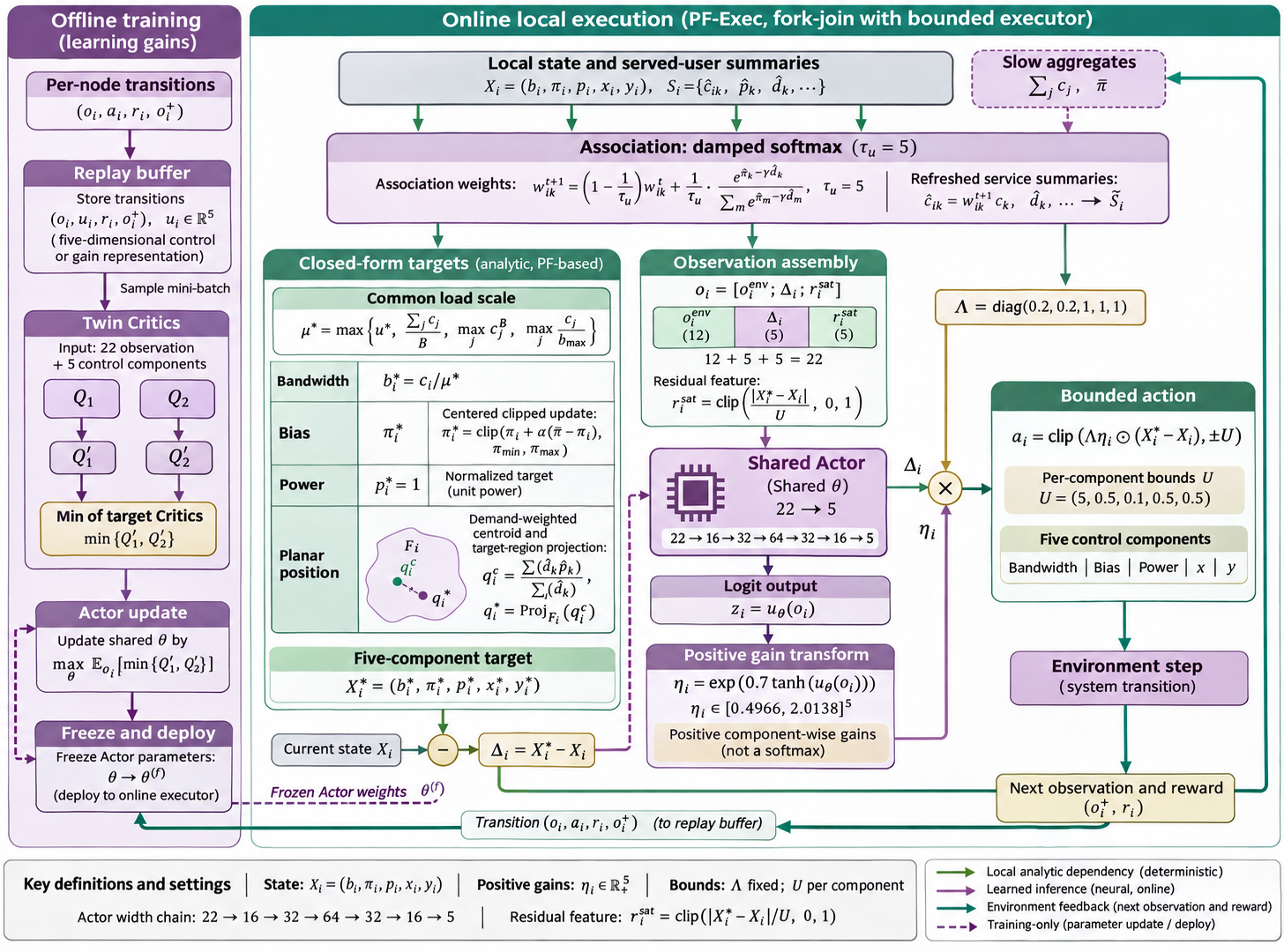}
\caption{Computation and training flow. The direction decider returns the closed-form targets and the direction package from the local state, the executor, chosen among constant speed, fixed step and learned gain, composes the direction and the residual distance into the five-dimensional action, and the environment advances on the channel, the association, the crowd and the physics before the state is read again. The network of the learned executor takes twenty-two inputs and shares them across all scales.}
\label{fig:framework}
\end{figure}

\subsection{Training}

The executor is trained by TD3 \cite{r60}, as Algorithm~\ref{alg:train} states. Twin critics with a minimum suppress overestimation, target policy smoothing suppresses spikes, and the actor updates every second step. Transitions are stored per UAV, so one environment step contributes $n$ transitions, and the exploration noise is added in the logit domain so that the gain bound holds during exploration as well. Gradients are clipped at ten, and the budget is 1200 rounds of 256 steps, about three hundred thousand steps.

\begin{figure}[!t]
\centering
\includegraphics[width=\columnwidth]{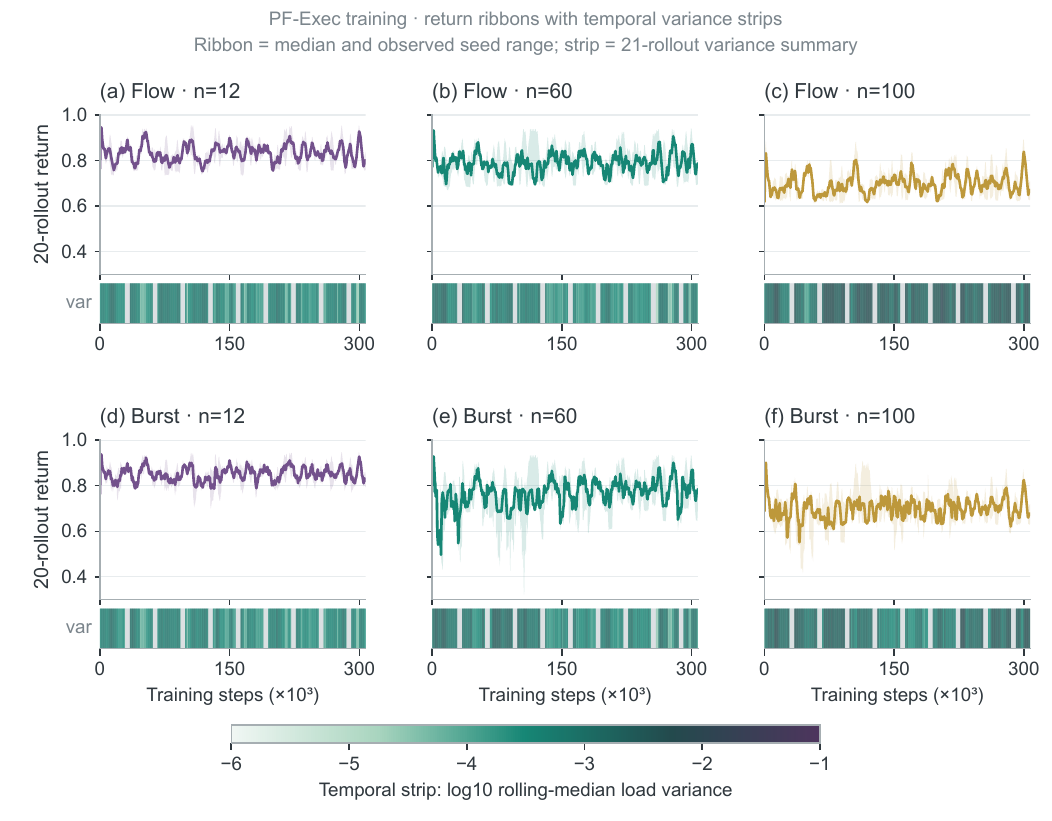}
\caption{Return band over training rounds, with the sliding mean on the vertical axis and the spread across seeds as a shaded band. The return starts from the pure potential starting point of $\eta\equiv1$, climbs, and enters the flat band that triggers the early stop.}
\label{fig:training}
\end{figure}

An early-stop rule watches two horizons. Training stops when the mean return of the last ten rounds sits within 0.5 per cent of the mean return of the last two hundred, which is a two-sided flat band, and a collapse does not trigger it. The rule is checked every 25 rounds together with a checkpoint. Fig.~\ref{fig:training} shows the return band over training. The return climbs from the pure potential starting point of $\eta\equiv1$, settles inside the flat band, and the band of seed spread narrows as training proceeds.

\begin{algorithm}[!t]
\caption{Executor training}
\label{alg:train}
\footnotesize
\begin{algorithmic}[1]
\STATE initialise actor with a zero output layer, twin critics and target copies
\FOR{round $=1$ to $1200$}
\FOR{step $=1$ to $256$}
\STATE run the direction decider for every UAV
\STATE add logit-domain exploration noise, form the gain and the action by (\ref{eq:gain}) and (\ref{eq:compose})
\STATE step the environment, store $n$ per-UAV transitions
\STATE update both critics on a batch of 64, update the actor every second step
\STATE soft-update the targets with rate 0.005
\ENDFOR
\IF{round is a multiple of 25}
\STATE evaluate the flat band over ten and two hundred rounds, checkpoint, stop when flat
\ENDIF
\ENDFOR
\end{algorithmic}
\end{algorithm}

\begin{table}[!t]
\centering
\caption{Executor network and training settings, identical at every scale.}
\label{tab:training}
\footnotesize
\begin{tabular}{@{}ll@{}}
\toprule
Item & Setting \\
\midrule
Observation & 22: 12 env, 5 direction, 5 residual \\
Hidden widths & 16, 32, 64, 32, 16 \\
Output & 5 positive gains, zero-initialised last layer \\
Gain map & $\eta=\exp(\pm0.7\tanh(\cdot))$ \\
Gain range & $[0.497,2.014]$ \\
Relaxation on $b,\pi$ & $0.2$ \\
Action clip & physical single-step bound \\
Algorithm & TD3, twin critics, delayed actor \\
Discount, batch & 0.99, 64 \\
Learning rate, soft update & $10^{-3}$, 0.005 \\
Exploration noise & 0.3 in the logit domain \\
Gradient clip & 10.0 \\
Budget & 1200 rounds of 256 steps \\
Early stop & 0.5 per cent band, every 25 rounds \\
\bottomrule
\end{tabular}
\end{table}

Table~\ref{tab:training} lists the settings, and none of them depends on the fleet size. The observation vector is the twelve environmental scalars of Section~II, the five components of the direction that the decider emits and the five normalised residuals between the target and the state, so the input dimension is set by the action dimension only. The last layer starts at zero, which makes the initial gain equal to one in every dimension and starts training from the pure potential policy, and the bounded exponential keeps the gain positive and inside $[0.497,2.014]$ throughout training and deployment.

\subsection{Reward and Complexity}

The reward of a UAV is
\begin{equation}
r_i=0.45\,\mathrm{bal}_i+0.35\,\mathrm{load}_i+0.20\,\mathrm{eff}_i,
\label{eq:reward}
\end{equation}
with the neighbourhood balance
\begin{equation}
\mathrm{bal}_i=\frac{1}{1+\sum_{j\in\mathcal N(i)}\bigl(\hat u_i-\hat u_j\bigr)^2/(\deg_i+1)},
\end{equation}
over the 250 m neighbourhood, which equals one for an isolated UAV. The load term is a quadratic platform around the anchor for $\hat u_i\le1$, with a width coefficient of 0.141, and a linear downhill of unit slope with a floor of $-3$ above one. The efficiency term is $\mathrm{eff}_i=\mathrm{clip}(T_i/\bar b/12,0,1)$, where the normalising ceiling of 12 Mbps per MHz sits above the spectral efficiency reachable in the normal load regime, about 6.7 b/s/Hz at 20 dB, so the term scales without reordering. The reward lives on the whole real line, and a negative value carries the depth of overload.

The complexity is local. The decider spends work proportional to the number of users a UAV serves, fleet coupling is two broadcast scalars, the sum of demand scales $\sum_jc_j$ and the mean price $\bar\pi$, with the total budget $B$ a public fixed parameter, and the user side of the association needs two further scalars, the price and the rate margin, and the normalisation $\sum_ix_{ik}=1$ is taken there, so no logit crosses the fleet, for a total of $O(n)$ independent of the number of users. The network input and output do not depend on the fleet size, and the whole system costs $O(nK)$ per step with a constant that does not grow with $n$.
\section{Experimental Setup}

\subsection{Scale Ladder}

The scale ladder keeps the density invariant, as Table~\ref{tab:scale} shows. The regional host base station is assumed to provide ample access-channel capacity, and the side of the field grows as $\sqrt{n/12}$, the number of users and the total bandwidth grow as $n/12$, and the bandwidth share of a UAV stays at 8.33 MHz at every scale, with fifty users per UAV on average. Three seeds are trained and evaluated independently and every grid is reported, with the minimum and the range made explicit.

\begin{table}[!t]
\centering
\caption{Scale ladder under the density-invariant convention.}
\label{tab:scale}
\footnotesize
\setlength{\tabcolsep}{4pt}
\begin{tabular}{@{}rrrrr@{}}
\toprule
$n$ & field (m) & users & total bandwidth & share per UAV \\
\midrule
3 & $300\times225$ & 150 & 25 MHz & 8.33 MHz \\
12 & $600\times450$ & 600 & 100 MHz & 8.33 MHz \\
30 & $949\times712$ & 1500 & 250 MHz & 8.33 MHz \\
60 & $1342\times1006$ & 3000 & 500 MHz & 8.33 MHz \\
100 & $1732\times1299$ & 5000 & 833 MHz & 8.33 MHz \\
\bottomrule
\end{tabular}
\end{table}

\subsection{Comparison Methods}

Ten comparison methods share one environment and one evaluation flow, namely eight external baselines and the two constant-speed and fixed-step executors that share the direction of PF-Exec, as Table~\ref{tab:methods} lists. Four of the external baselines need no training. The random policy samples the five-dimensional action uniformly. The fixed-parameter policy holds position and splits bandwidth evenly. The greedy heuristic sends each UAV to the demand-weighted centroid of the busiest quarter of the users and splits bandwidth evenly. The round-robin policy rotates users across UAVs on a 99 to 1 duty cycle.

Four further methods are the published baselines, each reproduced in its own native setting as Table~\ref{tab:native} lists, and each also adapted to the main setting. The block-coordinate method of Shen et al.\ \cite{r34} iterates bandwidth and position with a period of 400 and a slot of 20, in a centralised manner. The particle-swarm method of Jiang et al.\ \cite{r43} places UAVs with a swarm of twelve particles, forty iterations and a fairness fitness of exponent two. The DDPG deployment agent of Cheng et al.\ \cite{r53} keeps the original network and reward and trains from scratch, with OU exploration at $\theta=0.15$ and $\sigma=0.2$. The masked SDQN agent of Rizvi et al.\ \cite{r54} uses an 81-cell action, eight moves and hover by three power levels by three bandwidth levels, with action masking, and trains from scratch.

The last two of the eleven rows are the constant-speed and fixed-step executors that share the direction of PF-Exec.

\begin{table*}[!t]
\centering
\caption{Comparison methods, all sharing one environment and one evaluation flow.}
\label{tab:methods}
\footnotesize
\setlength{\tabcolsep}{8pt}
\begin{tabular}{@{}lll@{}}
\toprule
Method & Type & Speed or solver \\
\midrule
PF-Exec & proposed & learned gains \\
fixed-step & proposed executor & bandwidth 0.2, price 0.05, rest maximal \\
constant-speed & proposed executor & always maximal \\
greedy & untrained & top quarter demand centroid, even bandwidth \\
fixed-parameter & untrained & fixed position, even bandwidth \\
round-robin & untrained & 99 to 1 rotation \\
random & untrained & uniform sampling \\
Shen \cite{r34} & external, block coordinate & period 400, slot 20, centralised \\
Jiang \cite{r43} & external, particle swarm & 12 particles, 40 iterations, fairness exponent 2 \\
DDPG \cite{r53} & external, end-to-end & original network and reward, OU noise \\
masked SDQN \cite{r54} & external, end-to-end & 81 cells, 8 moves, 3 power, 3 bandwidth \\
\bottomrule
\end{tabular}
\end{table*}

\begin{table}[!t]
\centering
\caption{Native settings of the published baselines, every parameter taken from the source.}
\label{tab:native}
\footnotesize
\setlength{\tabcolsep}{4pt}
\begin{tabular}{@{}llllp{0.24\columnwidth}@{}}
\toprule
Source & UAVs & users & field & bandwidth, mobility \\
\midrule
Cheng \cite{r53} & 7 & 50 & $1000\times1000$ m & $20$ MHz per UAV, static \\
Jiang \cite{r43} & 5 & 250 & $500\times500$ m & 180 kHz per user, static \\
Rizvi \cite{r54} & 3 & 6 & $1000\times1000$ m & 15 kHz per UAV, linear 10 m/s \\
Shen \cite{r34} & 3 & 20 & $2000\times2000$ m & 10 MHz, static \\
\bottomrule
\end{tabular}
\end{table}

\noindent\textbf{Comparison methods are tuned to their best form and the proposed method is not tuned at all.} Every comparison obtains the best form it can take on this task. The four untrained methods take the strongest form of their rule. The two external optimisers take the best configuration after tuning, with the period-slot pair of Shen at 400 and 20 and the swarm of Jiang at 12 particles and 40 iterations, which are the tuned sizes that lose nothing in capability. The two learned baselines are trained from scratch under their original recipes and then tuned. The fixed-step executor is itself the product of tuning, with its four constants optimised grid by grid, which makes it a strong fixed-gain counterpart of the proposed executor. The three constants of the proposed method, the anchor of 0.8, the knee of 3 and the gain bound of $[0.50,2.01]$, are fixed once in advance and used unchanged across every scale, both scenarios, all seeds and the four native settings. The comparison therefore reads as an untuned proposed method against comparison methods tuned to their best, and the bias of that comparison runs against the proposed method.

\subsection{Evaluation Protocol}

Every run is deterministic in execution and is scored on three horizons, collected from one continuous trajectory extended without reset, as Table~\ref{tab:records} summarises for the evaluation scale. The horizons are 2000, 3000 and 4000 steps. The front window is the mean over steps 700 to 999 and the end window is the mean over the last three hundred steps. The indicators are the fairness index, the load variance together with a truncated variance at a load cap of five, the mean load, the number of overloaded UAVs, the number of dead UAVs, the coverage ratio of users whose best-link signal-to-interference-plus-noise ratio reaches 5 dB, the total throughput, the flight energy and the cumulative travel. Every run contributes one row of eleven indicators to the main table.

The records fall into three blocks, as Table~\ref{tab:records} separates them. The native block holds sixty records, one for each published baseline in its own setting, with the proposed method run on the same setting and the away line of that baseline on the main ladder. The main block holds 452 records over the five scales, the two crowd scenarios and three seeds, covering the proposed method, the constant-speed and fixed-step executors and the four untrained heuristics on every grid and the four trained baselines on the grids their ported setting supports. The degradation block holds 90 records on the same grids as the main block at the two largest scales, with decision-side information only. The three horizons of one grid share one execution, so the record count counts scored horizons and not independent trajectories.

A real information-degradation line repeats the main configurations with decision-side information only. The decision input refreshes user positions every 0.5 s with ten metres of Gaussian noise and holds them between refreshes, while the environment advances on true positions throughout. The paired difference against the ideal-information line measures the information the decision chain needs against the information the environment provides.

\begin{table}[!t]
\centering
\caption{Composition of the evaluation records, 602 in total.}
\label{tab:records}
\footnotesize
\setlength{\tabcolsep}{4pt}
\begin{tabular}{@{}p{3.0cm}p{4.2cm}r@{}}
\toprule
Block & Content & records \\
\midrule
native settings & four settings, proposed and original method, seeds and step horizons & 60 \\
main comparison & two scenarios, five scales, eleven methods, seeds and horizons & 452 \\
degradation line & two scenarios, five scales, three seeds, three horizons & 90 \\
\bottomrule
\end{tabular}
\end{table}

Records are counted per scored horizon, not per simulated trajectory. The three horizons of one configuration are read from one continuous trajectory extended without reset, so a block of records descends from fewer executions than its count suggests, the horizons of one cell are not independent, and the comparison across horizons within a cell is paired. The native block covers four published settings, each with the proposed method and with the original method of that setting, three seeds and the three horizons. The main block covers two scenarios and five scales over eleven methods, with three seeds for the proposed method and for the fixed-step executor and one seed per grid for each external baseline, again at the three horizons. The degradation line covers two scenarios and five scales at three seeds and the three horizons. The block counts are 60, 452 and 90, and they are taken from the record index of the runs; they do not follow from a product of settings, methods, seeds and horizons, because the seeds and horizons of one configuration are shared and the block sizes follow the index itself.

The indicators are defined so that a value can be read without reference to the others. The fairness index is the Jain index on the raw load, reported as it comes from the measurement without smoothing. The load variance is the variance of the same raw load across the fleet, and the truncated variant from Appendix~B caps the load at five so that a UAV carrying many times its share does not dominate the sum. A dead UAV is one whose load exceeds the same threshold, which is the operational definition of a UAV that can no longer meet the demand assigned to it. Coverage is the fraction of users whose served rate meets their own demand, which is the coverage indicator of the underlying demand model and needs no further threshold. Mean throughput is the sum of the served rates, and flight energy and cumulative travel are integrated over the run on the physical trajectory. Every indicator is computed from the same record, so a table entry and a curve in the figure panel read the same field.

Three conventions keep the comparison honest. The best form of each method is used, which for the external optimisers means the tuned configuration named in Table~\ref{tab:native} and for the heuristics the strongest version of the rule. The proposed method is not tuned on the comparison grids, so its parameters come from one training run on the main scenario and are carried to every grid unchanged. And every method is evaluated through the same clip and the same physical step bound, so no method is given a larger action than another.

\section{Results}

\subsection{Main Scenarios Across Scale}

\begin{figure}[!t]
\centering
\includegraphics[width=\columnwidth]{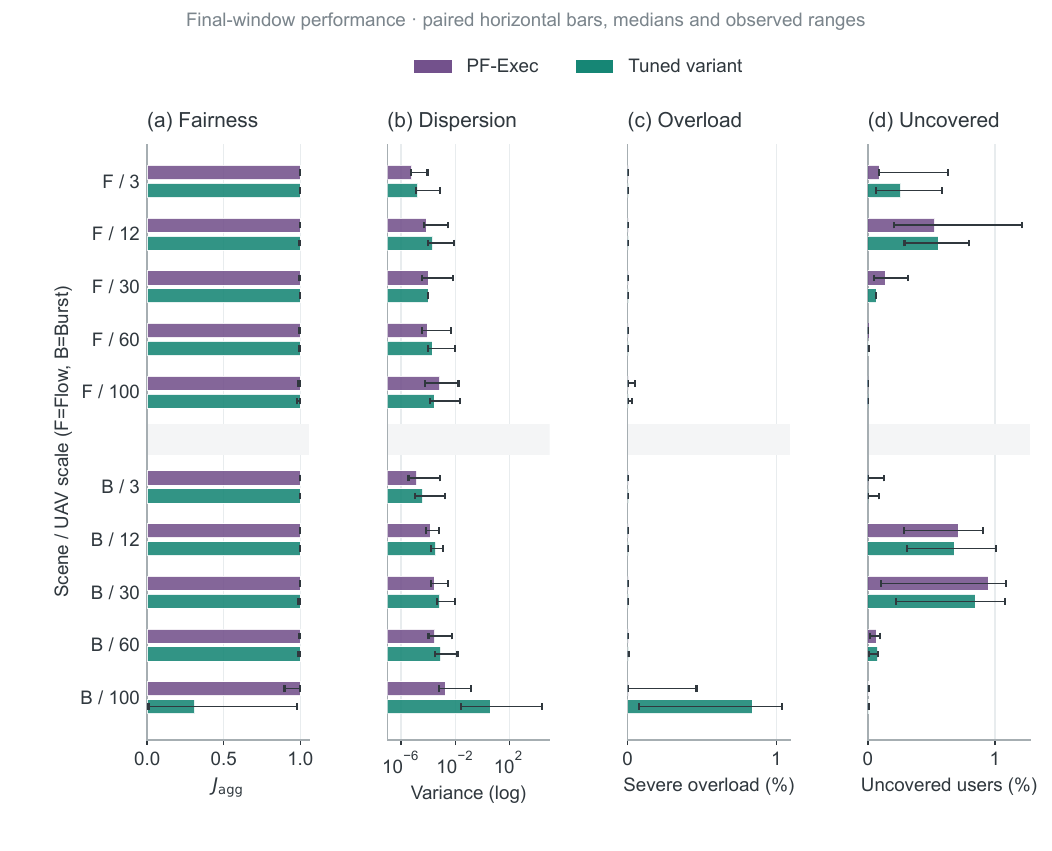}
\caption{Paired panels of the four indicators across the scale ladder, with the proposed method against the fixed-step executor. Seeds are marked separately. Fairness, dead count, coverage and load variance keep their ordering along the whole axis.}
\label{fig:dashboard}
\end{figure}

The main scenarios are compared at the longest horizon, where Table~\ref{tab:main} reports the end window and Fig.~\ref{fig:dashboard} pairs the four indicators across scale. The horizon is 4000 steps. PF-Exec holds a fairness index of at least 0.994 on twenty-eight of thirty grids, a dead count of at most 0.463 in window mean, at one hundred UAVs under burst load on seed~2, and a coverage ratio of at least 0.988 everywhere.

\begin{figure}[!t]
\centering
\includegraphics[width=\columnwidth]{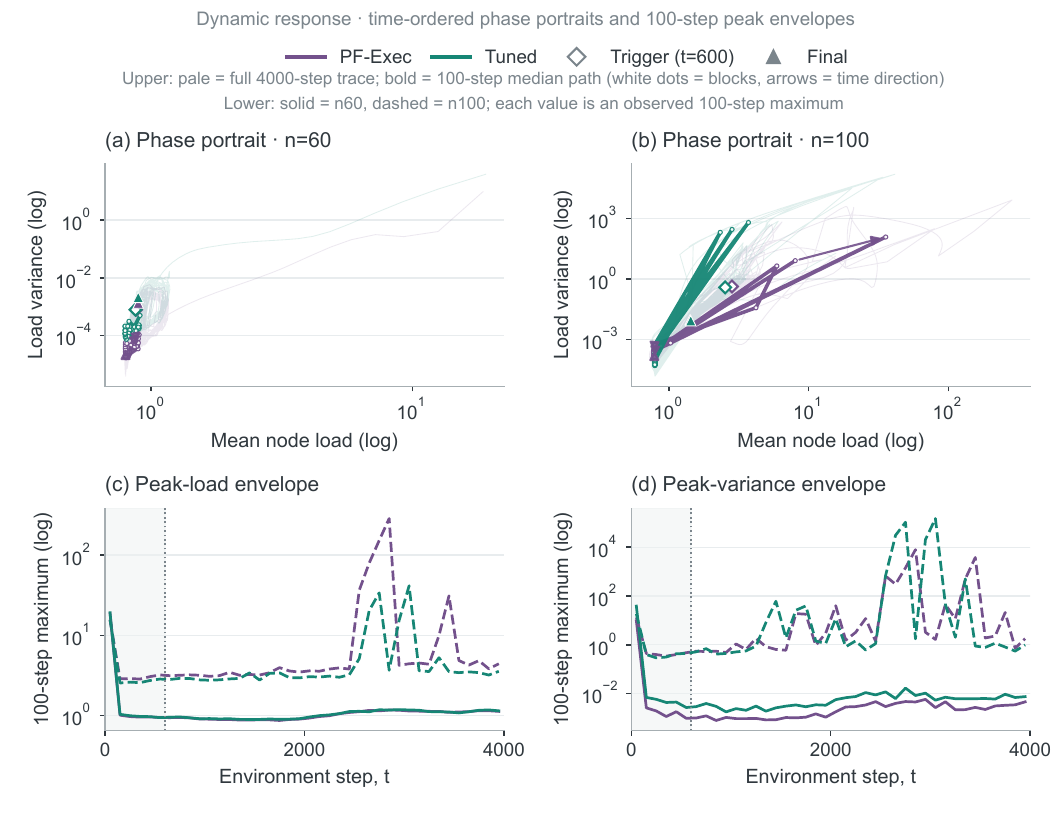}
\caption{Phase diagram under burst load, with mean load and the logarithm of load variance on the two axes, together with the envelope of the hundred-step peaks after the gathering at step 600. The trajectory slides along the balanced manifold and returns, and the envelope decays within a hundred steps.}
\label{fig:phase}
\end{figure}

Under flow the mean load rises gently from 0.80 to 1.06 along the scale axis, and on the regular grids the load variance of PF-Exec sits in the $10^{-5}$--$10^{-3}$ band. Under burst the mean load rises to the band from 0.80 to 1.21 after the gathering. The two one-hundred-UAV burst grids of seeds 0 and 1 are where the executors separate: the fixed-step executor loses both fairness and mean load there, with variance 3.88 and $2.52\times10^{4}$, while PF-Exec returns 0.9987 and 0.9995 with variance lower by a factor of seventeen than the greedy rule. The phase diagram of Fig.~\ref{fig:phase} traces the mechanism. After the gathering trigger the state slides along the balanced manifold and flows back, the variance lifts for a moment and returns to the anchor band, and the envelope of the hundred-step peaks decays within a hundred steps.

The highest-pressure setting separates the two executors. At one hundred UAVs under burst load, the fixed-step executor loses both fairness and mean load on seeds 0 and 1, with fairness 0.310 and 0.0089, mean load 1.32 and 15.05, variance 3.88 and $2.52\times10^{4}$, and a dead count of 0.84 and 1.04, while the same grids give 0.9987 and 0.9995 for PF-Exec. Seed~2 of that scale keeps the fixed-step executor at 0.976, and PF-Exec gives 0.897 there with a dead count of 0.463, so the advantage of the learned gain concentrates in the grids where aggregation pressure is highest. At one hundred UAVs under flow both executors stay healthy, at 0.9999, 0.9991 and 0.9846 for PF-Exec, which locates the split in the pressure of the gathering, since the scale alone leaves both executors healthy. The seed-to-seed spread of the fairness index runs from 0.002 to 0.10 across the grids.

The mechanism follows the step-size analysis of \cite{r67}. The ill-conditioning gain grows as the spectral sum shrinks or as the carried demand leaves the mid range, and both happen at once when the crowd gathers, so one fixed step cannot keep a single contraction condition across that range while the bounded truncated gain of PF-Exec keeps every dimension inside its cycle bound.

\begin{table*}[!t]
\centering
\caption{Main scenarios, end window at 4000 steps over three seeds. $J$ is the fairness index, $\bar u$ the mean load, var$_u$ the load variance, dead the window-mean count of dead UAVs and cov the coverage ratio. Ranges are over seeds.}
\label{tab:main}
\footnotesize
\setlength{\tabcolsep}{4.5pt}
\begin{tabular}{@{}llcccccc@{}}
\toprule
Scenario & $n$ & Method & $J$ & $\bar u$ & var$_u$ & dead & cov \\
\midrule
flow & 3 & PF-Exec & 0.9999--1.000 & 0.81--0.90 & $\le9.1\mathrm e{-5}$ & 0 & $\ge0.994$ \\
 &  & fixed-step & 0.999--1.000 & 0.81--0.89 & $\le7.3\mathrm e{-4}$ & 0 & $\ge0.994$ \\
 & 12 & PF-Exec & 0.9997--0.9999 & 0.80--0.94 & $\le3.0\mathrm e{-3}$ & 0 & $\ge0.988$ \\
 &  & fixed-step & 0.990--0.9998 & 0.80--0.93 & $\le8.6\mathrm e{-3}$ & 0 & $\ge0.992$ \\
 & 30 & PF-Exec & 0.994--0.9999 & 0.83--1.03 & $\le6.4\mathrm e{-3}$ & 0 & $\ge0.997$ \\
 &  & fixed-step & 0.9999 & 0.82 & $1.0\mathrm e{-4}$ & 0 & 0.999 \\
 & 60 & PF-Exec & 0.995--0.9999 & 0.81--1.00 & $\le5.0\mathrm e{-3}$ & 0 & 1.000 \\
 &  & fixed-step & 0.990--0.9998 & 0.81--0.99 & $\le9.7\mathrm e{-3}$ & 0 & $\ge0.9999$ \\
 & 100 & PF-Exec & 0.985--0.9999 & 0.87--1.06 & $\le1.8\mathrm e{-2}$ & $\le0.047$ & 1.000 \\
 &  & fixed-step & 0.981--0.9998 & 0.88--1.05 & $\le2.2\mathrm e{-2}$ & $\le0.030$ & 1.000 \\
\midrule
burst & 3 & PF-Exec & 0.999--1.000 & 0.80--0.86 & $\le7.1\mathrm e{-4}$ & 0 & $\ge0.999$ \\
 &  & fixed-step & 0.998--1.000 & 0.80--0.87 & $\le1.8\mathrm e{-3}$ & 0 & $\ge0.999$ \\
 & 12 & PF-Exec & 0.9991--0.9999 & 0.80--0.85 & $\le6.2\mathrm e{-4}$ & 0 & $\ge0.991$ \\
 &  & fixed-step & 0.998--0.9997 & 0.80--0.84 & $\le1.2\mathrm e{-3}$ & 0 & $\ge0.990$ \\
 & 30 & PF-Exec & 0.996--0.9998 & 0.86--0.93 & $\le3.1\mathrm e{-3}$ & 0 & $\ge0.989$ \\
 &  & fixed-step & 0.988--0.9994 & 0.87--0.95 & $\le1.0\mathrm e{-2}$ & 0 & $\ge0.989$ \\
 & 60 & PF-Exec & 0.995--0.9999 & 0.86--1.01 & $\le5.5\mathrm e{-3}$ & 0 & $\ge0.999$ \\
 &  & fixed-step & 0.985--0.9995 & 0.85--0.99 & $\le1.5\mathrm e{-2}$ & $\le0.007$ & $\ge0.999$ \\
 & 100 & PF-Exec & 0.897--0.9995 & 1.10--1.21 & $\le1.5\mathrm e{-1}$ & $\le0.463$ & 1.000 \\
 &  & fixed-step & 0.0089--0.976 & 1.06--15.1 & $\le2.5\mathrm e{4}$ & $\le1.04$ & $\ge0.9999$ \\
\bottomrule
\end{tabular}
\end{table*}

\subsection{Mechanism of the Collapse}

The collapse of the fixed-step executor has one cause, and the table shows where it bites. Write $\beta=\max_i\bigl\{1/S_i^{2},\ \kappa/(\rho_i+\varepsilon)^{2}\bigr\}$ for the ill-conditioning gain of the fleet, with $S_i$ the derivative of the access rate with respect to bandwidth at UAV $i$ and $\rho_i$ its carried demand. The sufficient step ceiling of a one-dimensional iteration of this shape scales as $\beta^{-1/2}$, so the admissible step shrinks when the spectral sum of a link falls and when the carried demand of a UAV leaves the mid range. Both happen during a gathering transition. Users concentrate into two exits, the access links of the loaded UAVs tighten while their demand jumps, and the fleet passes through a state in which the ceiling of the steepest dimension is far below the step that the flat dimensions want.

A fixed step carries a single value for every dimension and every state, so it must clear the smallest ceiling along the whole trajectory. Once the state crosses that point the per-dimension iteration leaves the geometric region, enters the bounded family of two-state cycles that the companion analysis characterises, and its per-coordinate error stops contracting at the anchor. The load profile inherits the oscillation, the mean load falls as the bandwidth shares swing, and at the two grids where the excursion is deepest the fairness index drops to 0.310 and 0.0089 while the mean load collapses with it. The physical clip keeps the action finite throughout, so the fixed-step executor never diverges. What it loses is the anchor, and the loss is permanent for as long as the pressure holds.

The learned executor carries a gain per dimension, computed from a local observation that contains the load, the residual and the direction, so the gain on a steep dimension can sit below the ceiling while the gain on a flat dimension stays near its maximum. The same mechanism explains the two sides of Table~\ref{tab:energy}. The learned gain does not travel farther, because it lowers the step where the map is steep; the flight energy and the travel distance stay within a fraction of a per cent of the fixed-step values, while the anchor is held at the grids where the fixed step loses it. The gain therefore buys survival at the anchor and not speed, which is the behaviour the reward is designed to produce and the behaviour the audit records in the per-dimension gain traces of Fig.~\ref{fig:phase}.

The mechanism also predicts where the advantage appears. The gap opens at the largest fleet under burst, where the demand per UAV jumps hardest and the gathering trigger compresses the crowd the most, and it closes at small fleets and under flow, where the ceiling stays above any fixed step that the comparison uses. Between the two ends the two executors track each other within the fairness band, which is what the scale sweep in Fig.~\ref{fig:dashboard} shows.

\subsection{Comparison Across the Fleet}

\begin{figure}[!t]
\centering
\includegraphics[width=\columnwidth]{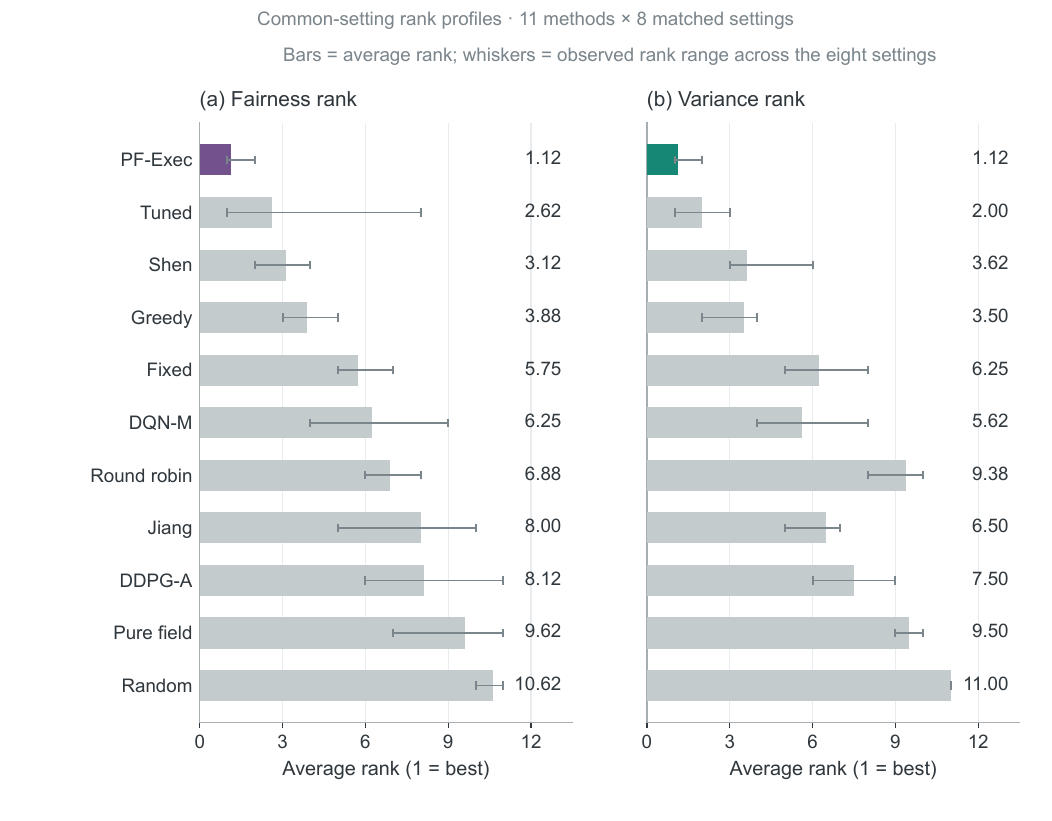}
\caption{Average rank over all common configurations, with rank spread as error bars. PF-Exec ranks first, the greedy heuristic and the block-coordinate method follow, and the random and constant-speed policies sit at the bottom.}
\label{fig:rank}
\end{figure}

The whole comparison system is evaluated on the longest horizon with one seed, where Table~\ref{tab:baseline} reports the end window at 4000 steps and Fig.~\ref{fig:rank} aggregates the average rank over all common configurations. Three observations stand out.

The contribution of direction and the contribution of speed separate cleanly. The constant-speed executor collapses everywhere and leaves a mean load of 176 at twelve UAVs under flow. The fixed-step executor sits close to PF-Exec at low and medium pressure and collapses only in the highest-pressure setting. PF-Exec survives every grid. The same direction handed to different speed laws spans the whole range from total collapse to full survival, which is the measured form of the layered argument.

Among the untrained heuristics the greedy rule is strongest, at 0.924, 0.967 and 0.976, with no dead UAV even at one hundred under burst. Sending each UAV to the centroid of its busiest users is itself a local balancing rule, which is why it survives. PF-Exec gives 0.9999, 0.9999 and 0.9987 on the same grids, and its load variance at one hundred UAVs under burst is lower by more than an order of magnitude, $1.9\times10^{-3}$ against $3.2\times10^{-2}$, a factor of seventeen, which comes from closed-form bandwidth anchoring against the even split of the heuristic.

The three contributions stay separable across the scale axis. At the smallest fleets the anchor is easy and every executor that keeps a positive gain holds a fairness index above 0.99, so the differences sit inside the seed spread and the figure shows the four indicators nearly overlapping. The separation opens with the fleet, because the demand that a gathering concentrates on one UAV scales with the number of users that the fused crowd delivers, while the bandwidth of that UAV scales with the fleet only through the density-invariant ladder. The middle of the ladder is where the fixed-step executor starts to drift from the anchor and where the two-state corridor of the companion analysis becomes visible in the variance trace. The top of the ladder separates the executors by survival, and the dashboard marks the two burst grids where the fixed-step line leaves the band altogether.

The external methods fail in different ways. The block-coordinate method reaches 0.998 and 0.997 at medium scale with no dead UAV, and its centralised closed-form bandwidth breaks down at one hundred UAVs under burst, with a mean load of $1.1\times10^{4}$ and the whole fleet dead. The particle-swarm method places UAVs offline and cannot follow a gathering, which gives 0.159 and 0.511 in the two burst grids. The DDPG and masked SDQN agents, trained from scratch, still lose the high-pressure burst grids, at 0.299 and 0.652, and the coverage of the masked agent falls to 0.60.

\begin{table*}[!t]
\centering
\caption{Comparison system at the end window of 4000 steps, seed~0. Dead counts are window means.}
\label{tab:baseline}
\footnotesize
\setlength{\tabcolsep}{4pt}
\begin{tabular}{@{}lccccccccc@{}}
\toprule
 & \multicolumn{3}{c}{flow, 12 UAVs} & \multicolumn{3}{c}{burst, 12 UAVs} & \multicolumn{3}{c}{burst, 100 UAVs} \\
\cmidrule(lr){2-4}\cmidrule(lr){5-7}\cmidrule(lr){8-10}
Method & $J$ & $\bar u$ & var$_u$ & $J$ & $\bar u$ & var$_u$ & $J$ & $\bar u$ & var$_u$ \\
\midrule
PF-Exec & 0.9999 & 0.80 & $4.8\mathrm e{-5}$ & 0.9999 & 0.80 & $6.7\mathrm e{-5}$ & 0.9987 & 1.21 & $1.9\mathrm e{-3}$ \\
fixed-step & 0.9998 & 0.80 & $1.1\mathrm e{-4}$ & 0.9997 & 0.80 & $1.7\mathrm e{-4}$ & 0.310 & 1.32 & 3.88 \\
constant-speed & 0.169 & 176 & $1.5\mathrm e{5}$ & 0.132 & 225 & $3.3\mathrm e{5}$ & 0.030 & $1.7\mathrm e{4}$ & $9.6\mathrm e{9}$ \\
greedy & 0.924 & 0.77 & $4.8\mathrm e{-2}$ & 0.967 & 0.68 & $1.6\mathrm e{-2}$ & 0.976 & 1.14 & $3.2\mathrm e{-2}$ \\
fixed-parameter & 0.871 & 0.79 & $9.2\mathrm e{-2}$ & 0.200 & 31.9 & $4.1\mathrm e{3}$ & 0.392 & $5.1\mathrm e{3}$ & $4.1\mathrm e{7}$ \\
round-robin & 0.770 & 63.9 & $1.2\mathrm e{3}$ & 0.175 & $2.6\mathrm e{3}$ & $3.1\mathrm e{7}$ & 0.364 & $4.3\mathrm e{5}$ & $3.2\mathrm e{11}$ \\
random & 0.100 & 991 & $8.8\mathrm e{6}$ & 0.029 & $1.9\mathrm e{4}$ & $1.2\mathrm e{10}$ & 0.042 & $1.1\mathrm e{7}$ & $2.5\mathrm e{15}$ \\
Shen \cite{r34} & 0.998 & 0.85 & $1.7\mathrm e{-3}$ & 0.997 & 0.81 & $1.8\mathrm e{-3}$ & 0.976 & $1.1\mathrm e{4}$ & $3.1\mathrm e{6}$ \\
Jiang \cite{r43} & 0.739 & 0.96 & $3.3\mathrm e{-1}$ & 0.159 & 7.9 & $3.3\mathrm e{2}$ & 0.511 & 699 & $4.7\mathrm e{5}$ \\
DDPG \cite{r53} & 0.616 & 1.04 & $6.7\mathrm e{-1}$ & 0.348 & 8.4 & $1.3\mathrm e{2}$ & 0.299 & $3.7\mathrm e{3}$ & $3.2\mathrm e{7}$ \\
masked SDQN \cite{r54} & 0.934 & 5.97 & 2.53 & 0.948 & 5.61 & 1.73 & 0.652 & 720 & $2.8\mathrm e{5}$ \\
\bottomrule
\end{tabular}
\end{table*}

\subsection{Native Settings of the Baselines}

\begin{figure}[!t]
\centering
\includegraphics[width=\columnwidth]{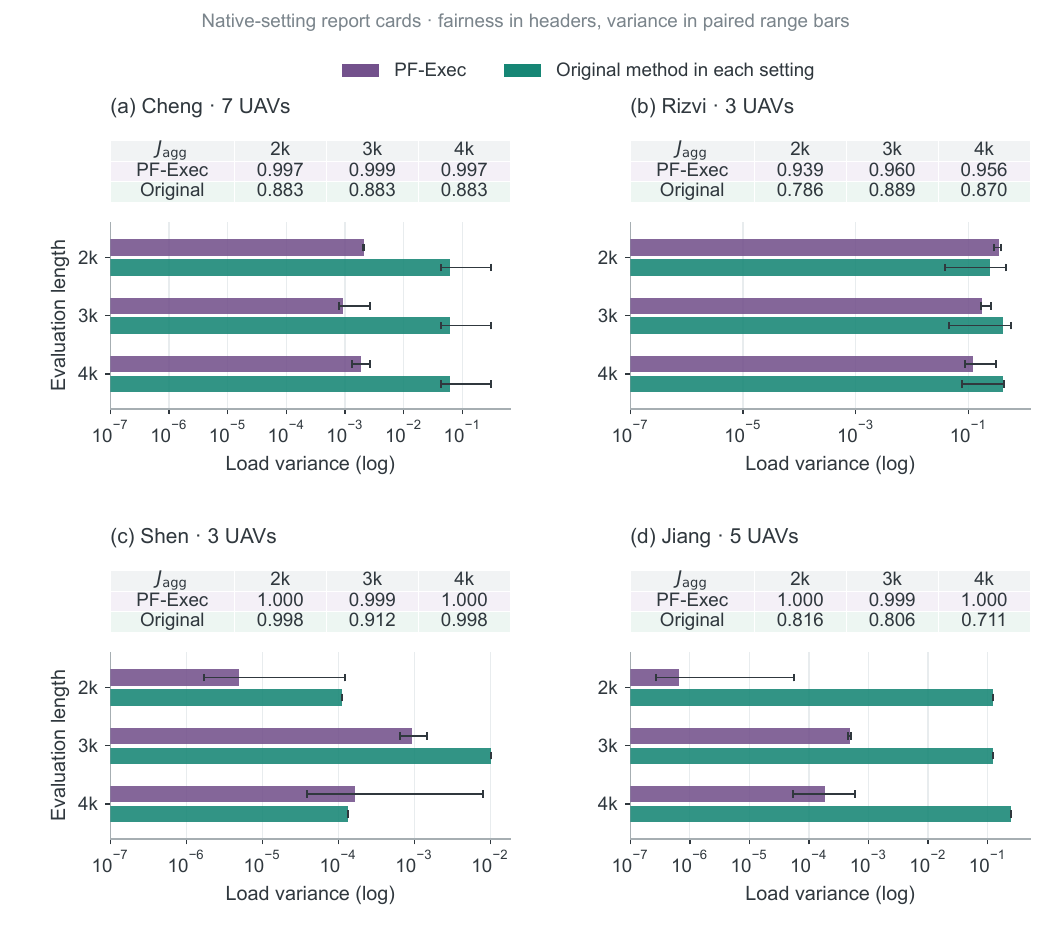}
\caption{Report cards of the four native settings. Each card gives the median fairness of the proposed method and of the original method at the 2000, 3000 and 4000 step horizons, with the median variance on a logarithmic axis.}
\label{fig:native}
\end{figure}

Fig.~\ref{fig:native} reports each published baseline in its own setting, with the median fairness over the three horizons and the median variance on a logarithmic axis. PF-Exec gives 0.997 against 0.883 for Cheng et al., whose reported interval runs from 0.752 to 0.943, 0.9997 against 0.711 for Jiang et al., 0.956 against 0.870 for Rizvi et al., whose interval runs from 0.850 to 0.924, and 0.9997 against 0.998 for Shen et al. No dead UAV appears in any setting.

The four settings differ in the quantity they were designed for, and the fairness index is a secondary indicator in two of them. The setting of Cheng et al.\ places seven UAVs over a thousand-metre field for fifty users on twenty megahertz, where coverage is the design target. The setting of Jiang et al.\ uses a hundred-and-eighty-kilohertz channel per user with a backhaul-dominated budget. The setting of Rizvi et al.\ is a low-altitude urban canyon with a narrow per-user channel. The setting of Shen et al.\ already reaches 0.998, and the proposed method reaches the same level, so no gain is claimed there. Each reproduction is run with its own published configuration and its own training budget, so the comparison keeps the internal validity of the original study.

PF-Exec matches or outperforms the reproduced methods in all four settings and improves clearly in three, using the network and hyperparameters of the main experiment and a separate training run. The transfer of the direction across settings comes from the dimensional self-consistency of the demand scale $c_i$, since the closed forms of Section~III adapt to the field size and the bandwidth regime without retuning. The reproductions of the original methods sit in the same direction as the published conclusions. The settings of Cheng et al.\ and Jiang et al.\ target coverage or delay, so inter-UAV fairness is not part of their objective and their lower fairness follows from that difference in objective. The setting of Shen et al.\ holds three UAVs and twenty users, where the balancing room is small, and the two methods nearly tie. The setting of Rizvi et al.\ runs at 15 kHz per UAV, the most extreme bandwidth scale in the set, and the bandwidth dimension of PF-Exec shrinks its step automatically because the single-step bound scales with the total bandwidth, which still holds a median of 0.956.

\subsection{Real Information Degradation}

\begin{figure}[!t]
\centering
\includegraphics[width=\columnwidth]{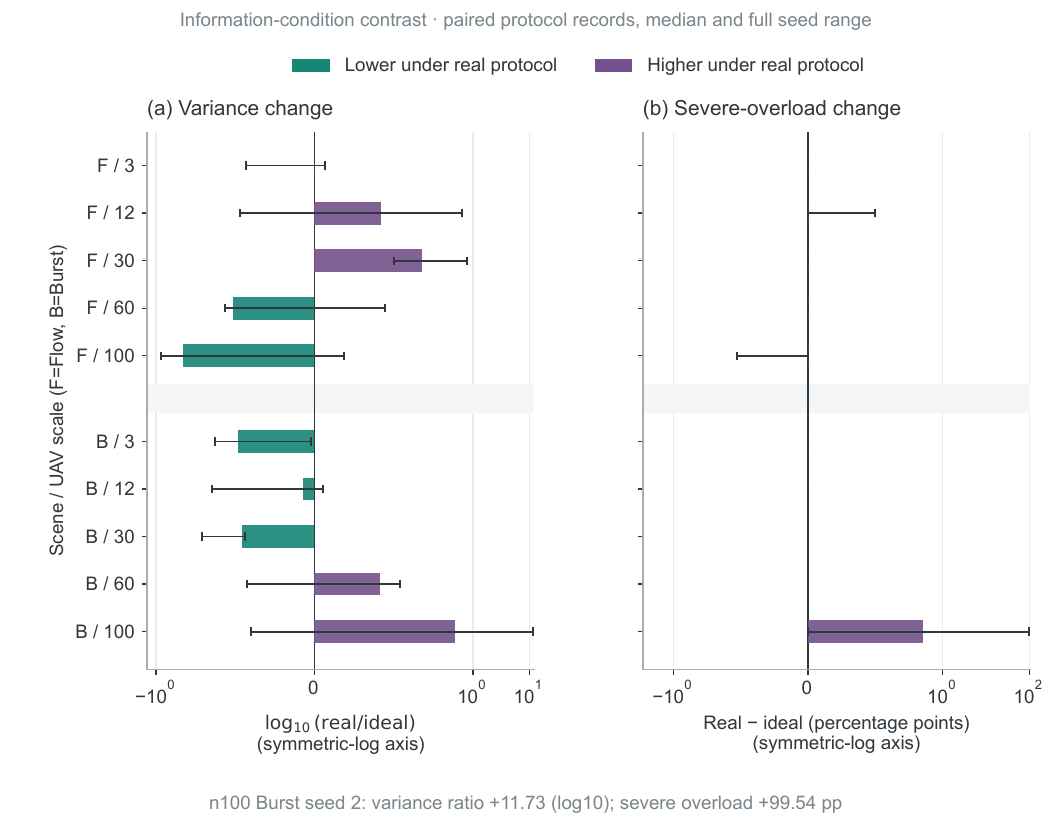}
\caption{Paired differences between the degradation line and the ideal-information line on identical grids. Differences concentrate below $10^{-3}$ and the largest, at one hundred UAVs under burst, is 0.002.}
\label{fig:protocol}
\end{figure}

The degradation line keeps the decision input on stale information while the environment advances on truth, so that user positions arrive as a snapshot carrying ten metres of noise and the fleet scalars refresh on the same period. The period is half a second, and seed~0 gives a fairness index of at least 0.9999 across the whole flow ladder, with variance at most $6.0\times10^{-5}$, no dead UAV and coverage at least 0.9985. Under burst the first four scales keep at least 0.9999 and the hundred-UAV scale gives 0.9967, with a dead count of 0.353 and a mean load of 1.32. Fig.~\ref{fig:protocol} pairs the degradation line against the ideal-information line grid by grid. The differences stay below $10^{-3}$ everywhere except at one hundred UAVs under burst, where the fairness difference reaches 0.002.

The mechanism is the low information frequency of the closed-form direction. The bandwidth target needs the local quantity and two broadcast scalars, the centroid needs only a half-second position sample, and no high-frequency global element sits in the decision chain, so the cost of degraded information is bounded and small.

\subsection{Energy and Travel}

Flight energy is almost identical between PF-Exec and the fixed-step executor, since the position dimension of the two shares one centroid target and the physical trajectories nearly coincide, while the learned gain acts on the resource dimensions and on the recovery phase. Table~\ref{tab:energy} reports flight energy, cumulative travel and total throughput over the full 4000 steps on seed~0, and balance carries no measurable flight-energy cost.

Under burst at one hundred UAVs throughput is $5.98\times10^{3}$ Mbps against $8.73\times10^{3}$ Mbps under flow. The crowd squeezes into two exits and access tightens on every UAV in the gathering, so the throughput of the scene falls for any policy that keeps the fleet alive. The measured value is what the scene allows, and PF-Exec reaches it with every UAV still serving.

\begin{table}[!t]
\centering
\caption{Flight energy, travel and throughput, seed~0 over the full 4000 steps.}
\label{tab:energy}
\footnotesize
\setlength{\tabcolsep}{4pt}
\begin{tabular}{@{}llrrr@{}}
\toprule
Scenario & Method & $e_{\mathrm{fly}}$ (J) & travel (m) & throughput (Mbps) \\
\midrule
flow, 12 & PF-Exec & $4.623\mathrm e{6}$ & 7490 & 1068 \\
 & fixed-step & $4.623\mathrm e{6}$ & 7450 & 1076 \\
flow, 100 & PF-Exec & $3.853\mathrm e{7}$ & $6.76\mathrm e{4}$ & 8728 \\
 & fixed-step & $3.854\mathrm e{7}$ & $6.78\mathrm e{4}$ & 8693 \\
burst, 12 & PF-Exec & $4.617\mathrm e{6}$ & 5300 & 1036 \\
 & fixed-step & $4.617\mathrm e{6}$ & 5403 & 1042 \\
burst, 100 & PF-Exec & $3.848\mathrm e{7}$ & $7.68\mathrm e{4}$ & 5979 \\
 & fixed-step & $3.849\mathrm{e}{7}$ & $7.72\mathrm e{4}$ & 5960 \\
\bottomrule
\end{tabular}
\end{table}

\section{Conclusion}

This paper splits relay load balancing into a direction and a speed. The anchored potential direction decider returns closed-form bandwidth, price, power and position targets from the local state, and the per-dimension gain executor converts those targets into actions under a hard bound that starts as the pure potential policy. The two layers form PF-Exec.

Across 602 evaluation records spanning three to one hundred UAVs, PF-Exec maintains a fairness index of at least 0.994 on twenty-eight of the thirty principal grids and avoids the collapse mode exhibited by fixed-step execution under severe demand aggregation. In the two 100-UAV burst grids where fixed-step fairness falls to 0.310 and 0.0089, PF-Exec restores near-unity fairness at 0.9987 and 0.9995. Against the strongest surviving heuristic on their common 100-UAV burst grid, PF-Exec raises fairness from 0.976 to 0.9987 and reduces load variance by a factor of seventeen. PF-Exec also matches or outperforms the reproduced methods across four native baseline settings and preserves high fairness under stale and noisy position information. The same three design constants are used across all scales and scenarios, while the reported paired comparisons show nearly identical flight energy between the learned and fixed-step executors. These results establish PF-Exec as a robust execution framework for large-scale UAV relay load balancing, combining closed-form recovery directions with learned gains to prevent execution collapse under severe demand shocks.

\appendices

\section{Homogeneity of the Load and the Closed-Form Targets}
\label{app:closedform}

Write the access rate of UAV $i$ as the sum over its served users of $x_{ik}R_{ik}$. With the bandwidth split of \eqref{eq:load}, the per-user share is $b_{ik}=x_{ik}d_kb_i/\sum_jx_{jk}d_j$, and every user under UAV $i$ sees the same spectral efficiency $r_{ik}$, so the access rate is linear in the bandwidth of the UAV,
\begin{equation}
R_i^{\mathrm A}=b_iS_i,\qquad
S_i=\sum_k\frac{x_{ik}^{2}d_k\,r_{ik}}{\rho_i},
\label{eq:appSi}
\end{equation}
with $\rho_i=\sum_kx_{ik}d_k$. The load per unit bandwidth follows as
\begin{equation}
c_i=\frac{\rho_i}{S_i}=u_ib_i,
\label{eq:appci}
\end{equation}
which does not depend on $b_i$. This is the homogeneity used in Section~III, and it is a property of the split itself, independent of the channel, since any split that makes the per-user share proportional to $x_{ik}d_k$ produces the same cancellation.

The load-equalising bandwidth target is the solution of $u_i=u^{\ast}$ for every UAV, that is $b_i^{\ast}=c_i/u^{\ast}$. If the sum exceeds the budget $B$, proportional rescaling multiplies every coordinate by the same factor and the target becomes
\begin{equation}
b_i^{+}=\begin{cases} b_i^{\ast}, & \sum_j b_j^{\ast}\le B,\\[2pt] \dfrac{B\,c_i}{\sum_jc_j}, & \sum_j b_j^{\ast}>B,\end{cases}
\label{eq:appscale}
\end{equation}
so the target stays proportional to the demand scale and, in the budget-binding branch, its common level rises to $\sum_jc_j/B$. The feasible level therefore satisfies
\begin{equation}
\mu_{\mathrm{low}}=\max\left\{\frac{\sum_ic_i}{B},\;\max_ic_i^{\mathrm B}\right\},
\label{eq:appmulow}
\end{equation}
where the second term is the backhaul-limited load of the worst UAV, since $u_i=\rho_i/\min\{b_iS_i,C_i^{\mathrm B}\}$ cannot fall below $\rho_i/C_i^{\mathrm B}$ when the backhaul binds.

\section{The Price Update in the Viewed-Load Coordinate}
\label{app:price}

With $\hat u_i=m(u_i)$ and $\Phi$ as in \eqref{eq:potential}, the derivative of $\Phi$ with respect to $\hat u_i$ is $n(\hat u_i-\bar{\hat u})$, which is the quantity the price dimension steers. The price update of \eqref{eq:price} is therefore a dual-inspired step in the view coordinate,
\begin{equation}
\pi_i^{+}=\pi_i+\bigl(m(u^{\ast})-m(u_i)\bigr)-\overline{\pi\,},
\label{eq:appprice}
\end{equation}
where the field removes the current mean price and then clips the result to the price range; the mean removal makes only relative prices matter, and the sum of prices after clipping is not asserted to vanish. The knee of $m$ at $u=3$ keeps the derivative positive on the whole bandwidth domain, $m'=1/(1+\max\{u-3,0\}/3)$, which is what preserves the ordering of the loads when the UAVs are deep in overload.

\section{Bounded Truncated Gain in One Dimension}
\label{app:gain}

Consider one dimension with target $\hat u^{\ast}$, view value $\hat u$ and effective gain $\eta\beta$, where $\beta$ collects the sensitivity of the view to the action. The composed recursion is
\begin{equation}
\hat u^{+}=\hat u+\operatorname{clip}\bigl(\eta\beta(\hat u^{\ast}-\hat u),\;-\Delta,\;\Delta\bigr),
\label{eq:appgain}
\end{equation}
with $\Delta$ the single-step bound in the view coordinate. As soon as $\eta\beta>2$ the truncated map leaves this regime. A step from $p$ with $|p|\ge\Delta/(\eta\beta)$ saturates and lands at $p-\Delta$, and the return step saturates as well whenever $p\le(1-1/(\eta\beta))\Delta$, so the map admits a family of two-cycles $\{p,\,p-\Delta\}$ with $p\in[\Delta/(\eta\beta),\,(1-1/(\eta\beta))\Delta]$, and that whole family, not a single attracting pair, is the invariant object. The largest amplitude over that family is $(1-1/(\eta\beta))\Delta<\Delta$, so an orbit stays inside the single-step slab, and at the upper end of the hard gain bound the amplitude is $0.503\Delta$. No value of the gain produces divergence in one dimension, and the three-dimensional coupling enters only through the targets, which is the statement used in Section~III.

\section{Two-Layer Shadowing}
\label{app:shadow}

The gain of a link is the sum of a spatial field and a time-varying term,
\begin{equation}
G_{ik}=C_0[k]-10\gamma\log_{10}(d_{ik}+50)+s_{ik},
\label{eq:appfit}
\end{equation}
where $C_0[k]$ is fitted per user by least squares over the baked grid and $\gamma$ is the median exponent over all users. The term $s_{ik}$ has a standard deviation drawn once per episode from $U[0,13.2]$ dB and is refreshed every hundred steps, which corresponds to the fifty metre decorrelation distance at the speeds of the fleet. The baked grid supplies the spatial correlation through the sum-of-sinusoids field of the channel generator, and the two layers are independent, so the online model keeps the spatial structure of the ray-based field while adding the variability that a static table cannot carry.

\section{Definitions of the Reported Indicators}
\label{app:indicators}

Every run reports the following quantities, evaluated on a window of steps. The load of UAV $i$ is $u_i=\rho_i/\max\{R_i^{\mathrm{avail}},10^{-9}\}$, where $\rho_i$ is the demand it serves and $R_i^{\mathrm{avail}}=\min\{R_i^{\mathrm A},C_i^{\mathrm B}\}$ is its end-to-end available rate; the floor keeps the ratio finite when nothing is served. In the narrowband setting the ratio of two near-zero quantities carries no information, so a UAV whose available rate falls below $10^{-6}$ Mbps is read as idle with $u_i=0$ and the load is capped at $3$, the knee of the view; the wideband settings use the plain ratio, where a nearly idle UAV is a normal part of the balance. A UAV is overloaded when $u_i>1$ and dead when $u_i>5$, and the dead test is the load threshold alone.

The fairness index is
\begin{equation}
J=\frac{\bar u^{2}}{\bar u^{2}+\operatorname{var}_u},
\label{eq:appJ}
\end{equation}
where $\bar u$ and $\operatorname{var}_u$ are the mean and the variance of the load over the UAVs in the window. A truncated variance is also reported, computed on the load capped at five, so that a single runaway UAV does not dominate the statistic.

The dead count is averaged over the window and is therefore reported as a real number, so that a single transient step stays distinguishable from a persistent failure. The overload count follows the same convention. Coverage is the fraction of users whose served rate meets their own demand, evaluated with the association of the current step and including the initial values of the episode. Throughput is the sum of the served rates over all users. Flight energy is the integral of the propeller power, $\sum_i\sum_tk_f\,\mathrm{rpm}^3\Delta t$, accumulated per UAV over the horizon, and travel is the sum of the displacement magnitudes.

The front window is the mean over steps 700 to 999, the end window is the mean over the last three hundred steps, and the three horizons of 2000, 3000 and 4000 steps are read from one continuous trajectory extended without reset, so the three horizons share a single trajectory and a single execution of every policy.

\section{Platform Implementation and Reproduction}
\label{app:platform}

The measurements of Section~V run on one implementation, and this appendix states the parts that the model of Section~II abbreviates.

\subsection{Channel Realisation}

The aerial channel follows TR~36.777 for the UMa-AV scenario, with the LoS probability as a function of height and ground distance and with the LoS and NLoS path losses taken as the larger of the free-space and the fitted expressions. Spatial correlation of the shadowing term comes from a sum-of-sinusoids generator with the UMa correlation distances, $\sigma=4$ dB over $37$ m on the LoS branch and $\sigma=6$ dB over $50$ m on the NLoS branch, and the LoS state of each user-UAV pair is drawn once per spatial realisation from a fixed seed. The gain table is baked offline at a horizontal resolution of $10$ m, which covers the field of every scale at the same grid density.

At run time each link keeps the statistical structure of the baked table and the small-scale term is averaged out, since the episode length of $200$ s is far above the coherence time of the fast fading. The dynamic part is the log-distance term, which is re-fitted per user by ordinary least squares on the loaded table as $G_{\mathrm{dB}}=C_0-10\gamma\log_{10}(d+50)$, with one global median slope and one intercept per user. The site-specific shadowing of a run is a Gaussian field whose standard deviation is drawn once per episode from $[0,13.2]$ dB and whose realisation refreshes every $100$ steps over a coherence distance of $50$ m. The upper end of that range is the residual of the fit against the measured table, so the run never assumes a cleaner channel than the table supports.

\subsection{Physics and Crowd}

The airframe reproduces a small quadrotor of $27$ g in a rigid-body simulator, stepped at $120$ Hz with a cascade position and attitude controller at $40$ Hz, so one environment step of $0.1$ s contains twelve physics ticks and four control updates. The trajectory interface exposes a horizontal displacement command whose single-step magnitude is clipped at $0.5$ m, which corresponds to $5$ m/s at the step length of the environment; the achieved speed over a run sits between $3.6$ and $4.4$ m/s because of the acceleration limit of the controller. Hovering height is held at $50$ m by the controller and is not part of the action.

The crowd is a social-force model with two groups, the fast group at $1.39$ m/s on odd indices and the slow group at $0.42$ m/s on even indices, walking inside a patrol radius of one standard deviation for the fast group and a third of it for the slow group. Under the burst scenario a gathering trigger fires at step $600$: two exogenous exits open, the crowd converges on them with a dwell radius of half a standard deviation, and the inflow speed rises to $2.78$ m/s. The user demand is redrawn per user from a truncated half-normal with mean scale $0.5$, standard deviation $0.15$ and support $[0.1,1]$, multiplied by the demand scale of the scale ladder, and it follows the crowd.

\subsection{Determinism and Artefacts}

An evaluation run is deterministic in execution: the seed fixes the crowd, the demand draw, the LoS states and the shadowing field, and the trained policies are evaluated without exploration noise. Every run writes one record with the eighteen headline indicators and one row per grid in the main table with forty-one columns, and the raw outputs behind them are archived with a content hash, $233$ files in total. The record identifier links the summary row to the raw file, so any entry of the tables of Section~V can be traced to the run that produced it. The theoretical audit of the companion analysis draws on thirty of these runs at the five scales with the pure field, and on its own summary tables, which is why its setting is stated separately from the main experiment.

\subsection{Budget of the Comparison}

The comparison system runs on one machine without a graphics device, since the physics and the crowd simulation carry the computational cost and the networks are small relative to them. One grid of $2000$ steps at one hundred UAVs costs minutes of wall time, and the trained baselines are retrained from scratch under their own published configurations on the same environment. The full record system of $602$ records therefore fits inside a single workstation of the kind the field already uses, which is what makes the sweep reproducible without a cluster.

\section{Rate, Load and the Closed-Form Target}
\label{app:rates}

This appendix derives the expressions that Section~III quotes, so that each closed form has a line of algebra behind it.

\subsection{Access Rate as Bandwidth Times a Spectral Coefficient}

With the demand-proportional split of A1, the carrier of the demand is $\rho_i=\sum_kx_{ik}d_k$ and the share of user $k$ inside the bandwidth of UAV $i$ is $b_i x_{ik}d_k/\rho_i$, so the access rate is the sum over associated users of the share times the spectral efficiency,
\begin{equation}
R_i^{\mathrm A}=b_i\,S_i,\qquad S_i=\sum_k \frac{x_{ik}^2 d_k\log_2(1+\gamma_{ik})}{\rho_i},
\label{eq:rateS}
\end{equation}
with $\gamma_{ik}$ the signal-to-interference-and-noise ratio of the link. The coefficient $S_i$ is the sum of the per-user efficiencies weighted by the squared association weight and the demand share, and it does not contain $b_i$. The carrier of the demand is
\begin{equation}
\rho_i=\sum_k x_{ik}d_k,
\label{eq:rho}
\end{equation}
which counts the demand that the association routes to UAV $i$, and the load is
\begin{equation}
u_i=\frac{\rho_i}{\min\{b_iS_i,\ C_i^{\mathrm B}\}},
\label{eq:uload}
\end{equation}
the ratio of carried demand to the rate that is actually available, where the backhaul cap enters through the minimum.

\subsection{First-Order Homogeneity}

In the access-limited region the minimum in \eqref{eq:uload} is the access term, and the load becomes
\begin{equation}
u_i=\frac{\rho_i}{b_iS_i}=\frac{c_i}{b_i},\qquad c_i=\frac{\rho_i}{S_i},
\label{eq:homogeneous}
\end{equation}
which is homogeneous of degree minus one in $b_i$: doubling the bandwidth of a UAV halves its load, and the demand scale $c_i$ carries no bandwidth at all. This is the property used in Lemma~1, and it holds for any split that makes the per-user share proportional to $x_{ik}d_k$, independently of the channel.

The equal-load bandwidth target follows by setting $c_i/b_i=u^{\ast}$ for every UAV,
\begin{equation}
b_i^{\ast}=\frac{c_i}{u^{\ast}},
\label{eq:btargetstar}
\end{equation}
and the target is a ratio of two scalars per UAV. Two constraints act on it. The budget requires $\sum_ib_i^{\ast}\le B$, which reads $u^{\ast}\ge\sum_ic_i/B$, and the backhaul requires $b_i^{\ast}S_i\le C_i^{\mathrm B}$, which reads $u^{\ast}\ge c_i^{\mathrm B}$ for the UAV with the largest backhaul demand scale. The feasible level is therefore
\begin{equation}
\mu_{\mathrm{low}}=\max\left\{\frac{\sum_ic_i}{B},\ \max_ic_i^{\mathrm B}\right\},
\label{eq:mulow}
\end{equation}
and any anchor above $\mu_{\mathrm{low}}$ is attainable.

\subsection{The Budget-Binding Branch}

When the budget binds, the target of \eqref{eq:btargetstar} is rescaled by the common factor $B/\sum_jb_j^{\ast}$, which gives
\begin{equation}
b_i^{+}=\frac{B\,c_i}{\sum_jc_j},
\label{eq:bbinding}
\end{equation}
and the common level of the load rises to $\sum_jc_j/B$. The rescaled target still equalises the load, since every UAV is multiplied by the same factor, so the profile stays flat and only its absolute level changes. This is what the experiment reports as the degradation of the anchor under pressure: the shape of the profile is preserved and the level moves, and the load view of Section~II records the level without changing the order of the UAVs.

\subsection{Price and Position Targets}

The price enters the decider through the subgradient of the potential with respect to the viewed load. With $\Phi=\frac12\sum_{i<j}(\hat u_i-\hat u_j)^2$ the derivative in one coordinate is $n(\hat u_i-\bar{\hat u})$, so the price target is the current price plus that difference, taken after subtracting the fleet mean and clipped to the admissible interval. The subtraction makes the update invariant to a common shift of all prices, which is the gauge that the environment fixes by announcing the mean price as a broadcast scalar.

The position part is separable. The position term of the potential is a weighted sum of squared distances between a UAV and its served users, which in each coordinate is a one-dimensional quadratic, and the constrained minimiser on the field rectangle is the coordinatewise clip of the weighted centroid. The position target is therefore the demand-weighted centroid of the served users, clipped to the field, and the direction on that part is the unit vector from the current position to the target.

\section{Ports of the Published Baselines}
\label{app:baselines}

The comparison mixes four ported learning agents, two ported optimisation methods and four untrained heuristics, and this appendix states what a port preserves and what it has to change.

\subsection{Learning Agents}

A ported agent keeps the architecture, the action parameterisation and the reward structure of its paper, and it is retrained from scratch on the common environment. Three things cannot be kept verbatim, and each port handles them the same way. The first is the fleet size, since the native settings of the four papers cover three to seven UAVs while the scale ladder runs to one hundred, so the input layer is defined by the observation vector of the common environment and the hidden widths follow the original ratio. The second is the action set, since the native papers act on position or on power alone while the common environment exposes five dimensions per UAV, so the action head is widened and the original constraint is applied to its own dimensions. The third is the reward, since the native rewards are defined on their own indicators and the common environment reports a fleet potential, so the shaping terms are expressed on the common indicators with the original weights rescaled to unit sum.

\subsection{Optimisation Methods}

The block-coordinate method keeps its two-loop structure: the price program is re-solved every four hundred steps and the bandwidth program together with the successive-convex-approximation trajectory every twenty steps, on the whole fleet, which is the centralised cost that the layered design removes. Both re-solutions use the current association and the current rates, and the method therefore carries the same information as the decider at a cost that grows with the fleet.

The particle-swarm method keeps its offline placement stage, which searches the positions of the UAVs against an alpha-fair effective-capacity fitness with twelve particles and forty iterations, and its online association step, which is a dual subgradient on the same fitness. Above twelve UAVs the particle count stays at twelve and the iteration count at forty, which is what makes the method degrade with the fleet: the search does not widen with the dimension it searches. The offline stage is also why the method cannot follow a gathering, since the positions it returns are fixed for the run.

\subsection{Untrained Heuristics}

The four heuristics carry no training and define the floor of the comparison. The random rule draws each action from the admissible box. The fixed rule splits bandwidth evenly and holds position. The greedy rule sends each UAV to the demand-weighted centroid of the busiest quarter of its users, which is itself a local balancing rule and, as Section~V reports, the strongest of the four. The round-robin rule rotates the served users on a duty cycle of one per cent, which spreads service but not load. Each heuristic is evaluated with the same clip and the same physical step bound as the proposed method.

\subsection{Reading the Comparison}

The comparison is read in three layers. Within the untrained layer the spread measures how much a local rule can achieve without any learning, and the greedy rule marks the level that a strong heuristic reaches. Within the fixed-gain layer the two companion executors locate the same direction at the two ends of the speed range, and the gap between them and PF-Exec measures what a state-dependent gain adds. Across the published layer the ports measure whether the layered design competes with specialised methods on their own terms, which is why each port is also evaluated in its native setting in Fig.~\ref{fig:native}.

\section{Implementation Notes on the Two Hops}
\label{app:implementation}

This appendix records how the quantities of the model are computed in the running system, because several of the closed forms are read at run time from measured values and the reading has to be stated to be reproducible.

\subsection{Backhaul and Access Rates}

The backhaul rate of UAV $i$ is written as $C_i^{\mathrm B}=\beta_i^{\mathrm B}\log_2(1+\gamma_i^{\mathrm B})$, where the coefficient $\beta_i^{\mathrm B}$ is the scheduled share of the backhaul and the signal-to-noise ratio is evaluated on the backhaul link with its own gain. The access rate is $R_i^{\mathrm A}=b_iS_i$, and the coefficient
\begin{equation*}
S_i=\sum_k x_{ik}^2d_k\kappa_k\Big/\rho_i,\qquad \rho_i=\sum_kx_{ik}d_k
\end{equation*}
is the bandwidth efficiency of UAV $i$ summed over the users it serves, with $d_k$ the demand of user $k$ and $\kappa_k$ the spectral efficiency of that user's link. The two quantities are evaluated on the same control step, so the serving hop of UAV $i$ is the minimum of the two and the load follows from the ratio without any interpolation.

The efficiency $k_k$ is not a model constant. At the start of a run the environment fits, per user, the slope and intercept of the path loss in decibels against the logarithm of the distance, from the shadowed gain samples that the channel model provides at ten-metre resolution. The fit is a two-parameter least squares, and the per-link deviation from the fit is carried as an additional loss term. The fit is refreshed every hundred steps and whenever a user has moved fifty metres, which keeps the fitted exponent close to the local one while avoiding a fit at every control step.

\subsection{The Load View at Run Time}

The view of Section~III is read from the state itself. The demand scale of UAV $i$ is $c_i=u_ib_i$, computed from the measured load and the measured bandwidth. In the access-limited region the product is invariant, so it can be read on any step without drift, and the drift observed over a run is at round-off. In the budget-bound region the same product is used and the anchor is the level that the rescaling assigns, which means the decider works with one expression in both regions.

The bandwidth target is then $b_i^\ast=c_i/\mu^\ast$ with $\mu^\ast$ the maximum of the budget-imposed level and the backhaul-imposed level, followed by the rescaling that puts the shares on the budget simplex. The rescaling is what makes the target feasible when the budget is tight, and it is also why the direction does not need the central solver: the shares are proportional to the local demand scales, so the only fleet-wide information is their sum.

\subsection{Order of the Updates}

One control step applies the blocks in a fixed order: the association updates with its damping, the bandwidth follows the closed-form target, the prices take one dual-inspired step and are clipped, the power multipliers take one step and are clipped, and the positions move towards the centroid with the physical bound applied last. The order is fixed so that the audit can recompute any state from its predecessor, and each block is idempotent in the sense that applying it twice at a fixed state gives the same state as applying it once.

The price step subtracts the mean price before clipping, which keeps the price vector from drifting as a whole and leaves only its dispersion to act on the allocation. The implementation does not claim that the sum of prices is zero after the clip, only that the update is invariant to a common shift before the clip.

\subsection{What the Decider Costs Per UAV}

Per control step and per UAV the decider evaluates one product for the demand scale, one division for the target, one sum over the served users for the position target, and one call to the executor network whose input and output dimensions are fixed. None of these depends on the number of UAVs, and the only quantities that couple the fleet are the sum of the demand scales and the mean price, both broadcast. A centralised block-coordinate method, by contrast, re-solves a price program and a bandwidth program over the whole fleet, and the cost of each re-solution grows with the number of UAVs and users together. The runtime difference is therefore structural and not an implementation detail, and it is the reason a fleet of one hundred UAVs stays within the same per-step budget as a fleet of three.

\vspace{6pt}

\noindent\textbf{Open source.} The implementation and the full experimental records are available at \url{https://github.com/jhzAliyy/space4-run}.

\end{document}